\documentclass[10pt,twocolumn,letterpaper]{article}

\usepackage{arxiv}

\usepackage[utf8]{inputenc} 
\usepackage[T1]{fontenc}    
\usepackage{url}            
\usepackage{booktabs}       
\usepackage{amsfonts}       
\usepackage{nicefrac}       
\usepackage{microtype}      
\usepackage{xcolor}         

\usepackage{graphicx}
\usepackage{booktabs}
\usepackage[utf8]{inputenc} 
\usepackage[T1]{fontenc}    
\usepackage{url}            
\usepackage{nicefrac}       
\usepackage{microtype}      
\usepackage{bbding}
\usepackage{pifont}
\usepackage{bbm}
\usepackage{bm} 
\usepackage{soul}
\usepackage{multirow}
\usepackage{multicol}
\usepackage{makecell}
\usepackage{framed}

\usepackage{amsmath}
\usepackage{amssymb}
\usepackage{mathtools}
\usepackage{amsthm}
\allowdisplaybreaks[4]

\usepackage{wrapfig}
\usepackage{caption}

\usepackage{amsmath}
\usepackage{amssymb}
\usepackage{mathtools}
\usepackage{amsthm}

\theoremstyle{plain}
\newtheorem{theorem}{Theorem}[section]
\newtheorem{proposition}[theorem]{Proposition}
\newtheorem{lemma}[theorem]{Lemma}
\newtheorem{corollary}[theorem]{Corollary}
\theoremstyle{definition}

\theoremstyle{remark}

\usepackage[textsize=tiny]{todonotes}

\usepackage{natbib}

\usepackage[pagebackref,breaklinks,colorlinks]{hyperref}

\usepackage{amsmath}
\usepackage{amssymb}
\usepackage{mathtools}
\usepackage{amsthm}
\usepackage{bbold}

\usepackage[capitalize,noabbrev]{cleveref}

\begin{document}
\theoremstyle{plain}
\theoremstyle{definition}
\theoremstyle{remark}

\title{Towards the Resistance of Neural Network Watermarking to Fine-tuning}

\author{
    \textbf{Ling Tang}\textsuperscript{1} \quad
    \textbf{Yuefeng Chen}\textsuperscript{2} \quad
    \textbf{Hui Xue}\textsuperscript{2} \quad
    \textbf{Quanshi Zhang}\textsuperscript{1}\thanks{Quanshi Zhang is the corresponding author. He is with the Department of Computer Science and Engineering, the John Hopcroft Center, at the Shanghai Jiao Tong University, China.} \\
    \textsuperscript{1}Shanghai Jiao Tong University \\
    \textsuperscript{2}Alibaba Group \\
}

\maketitle

\begin{abstract}
This paper proves a new watermarking method to embed the ownership information into a deep neural network (DNN), which is robust to fine-tuning. Specifically, we prove that when the input feature of a convolutional layer only contains low-frequency components, specific frequency components of the convolutional filter will not be changed by gradient descent during the fine-tuning process, where we propose a revised Fourier transform to extract frequency components from the convolutional filter.  Additionally, we also prove that these frequency components are equivariant to weight scaling and weight permutations. In this way, we design a watermark module to encode the watermark information to specific frequency components in a convolutional filter. Preliminary experiments demonstrate the effectiveness of our method.
\end{abstract}


\section{Introduction}
Watermarking techniques have long been used to protect the copyright of digital content, including images, videos, and audio \citep{nematollahi2017digital}. Recently, these techniques have been extended to protect the intellectual property of neural networks. Watermarking a neural network is usually conducted to implicitly embed the  ownership information into the neural network. In this way, if a neural network is stolen and further optimized, the ownership information embedded in the network can be used to verify its true origin. Previous studies usually embedded the ownership information in different ways. For example, \citet{wang2020watermarking} directly embedded the watermark into the network parameters. \citet{lukasdeep} used the classification results on a particular type of adversarial examples as the backdoor watermark. \citet{kirchenbauer2023watermark}  added a soft watermark to the generation result.

However, one of the core challenges of neural network watermarking is that most watermarking techniques cannot be resistant to the fine-tuning of the DNN. When network parameters are changed during the fine-tuning process, the watermark implicitly embedded in the parameters may also be overwritten.  

Although many studies \citep{uchida2017embedding} \citep{adi2018turning}\citep{liu2021watermarking} \citep{tan2023deep} \citep{zeng2023huref} have realized this problem and have tested the resistance of  their watermarks to fine-tuning, there is no theory designed towards the resistance of watermarking \emph{w.r.t.} fine-tuning, to the best of our knowledge.

The core challenge towards the resistance to fine-tuning is to explore an invariant term in the neural network to fine-tuning, \emph{e.g.}, certain network parameters or some properties of network parameters that are least affected during the fine-tuning process. Although \citet{zeng2023huref}, \citet{zhang2024reef}, and \citet{fernandez2024functional} have explored invariant terms \emph{w.r.t.} weight scaling and weight permutations for watermarking, the theoretically guaranteed invariant term to fine-tuning remains unsolved.

\begin{figure*}[t]
\vskip 0.2in
\begin{center}
\centerline{\includegraphics[width=\linewidth]{ 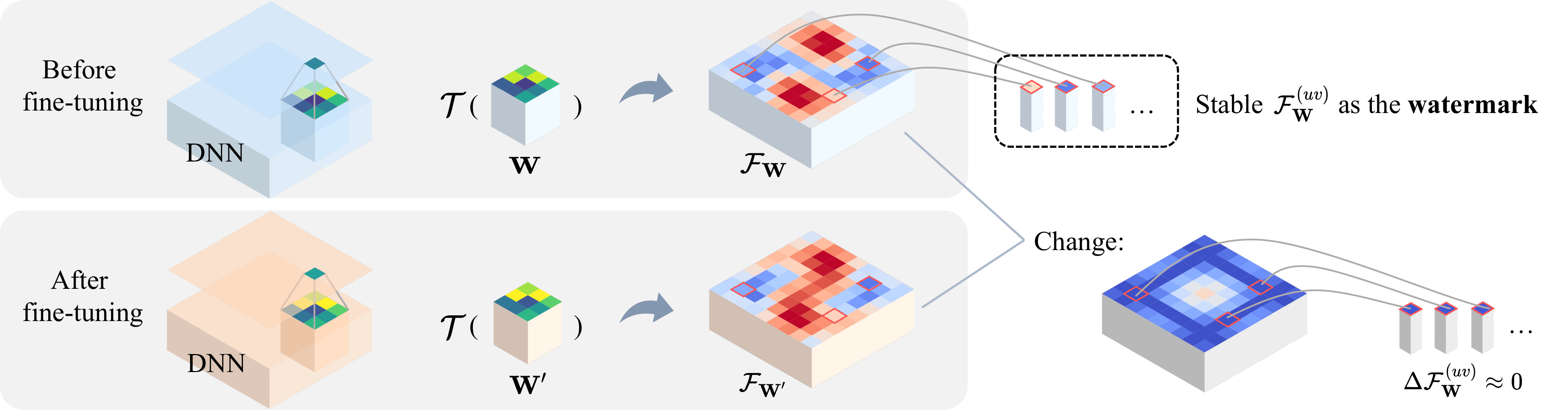}}
\caption{The framework of the proposed watermark. We prove that the specific frequency components\footref{fn:comp_filter} {\small $\mathcal{F}^{(uv)}_\mathbf{W}$}, which are obtained by conducting a revised discrete Fourier transform {\small $\mathcal{T}(\cdot)$} on the convolutional filter {\small $\mathbf{W}$}, keep stable in the training process. Thus, these specific frequency components {\small $\mathcal{F}^{(uv)}_\mathbf{W} $} are used as the robust watermark to fine-tuning. For clarity, we move low frequencies to the center of the spectrum map, and move high frequencies to corners of the spectrum map. Unless otherwise stated, in this paper, we visualize the frequency spectrum map in this manner.}
\label{fig_1}
\end{center}
\vskip -0.2in
\end{figure*}

In this study, we aim to discover and prove such an invariant term to fine-tuning. Specifically, as Figure \ref{fig_component} shows, \citet{tang2023defects} have found that the forward propagation through a convolutional layer {\small $ \mathbf{W}\otimes \mathbf{X} + b\cdot \mathbf{1}_{M\times N}$} can be reformulated as a specific vector multiplication between frequency components {\small $\mathcal{F}^{(uv)}_\mathbf{W} \cdot \mathcal{F}^{(uv)}_\mathbf{X} + \delta_{uv} MN b$} in the frequency domain, where  {\small $\mathcal{F}^{(uv)}_\mathbf{X}$} denotes the frequency component of the input feature {\small $\mathbf{X}$} at frequency {\small $(u,v)$}, which is extracted by conducting a discrete Fourier transform, {\small $\mathcal{F}^{(uv)}_\mathbf{W}$} denotes the frequency component\footnote{\label{fn:comp_filter} The frequency component {\small $\mathcal{F}^{(uv)}_\mathbf{W}$} of the convolutional filter is defined in Equation~(\ref{eq:com_filter}), which is extracted by applying a revised discrete Fourier transform on the convolutional filter {\small $\mathbf{W}$}. According to Theorem~\ref{th:forward}, the frequency component {\small $\mathcal{F}^{(uv)}_\mathbf{W}$} at frequency {\small $(u,v)$} represents the influence of the convolutional filter {\small $\mathbf{W}$} on the corresponding frequency component {\small $\mathcal{F}^{(uv)}_\mathbf{X}$} extracted from the input feature {\small $\mathbf{X}$}. } of the convolutional filter {\small $\mathbf{W}$}, and {\small $b$} is the bias term. 

Based on this, we prove that if the input feature  {\small $\mathbf{X}$} only contains the low-frequency components, \textit{then specific frequency components  of a convolutional filter {\small $\mathcal{F}^{(uv)}_{\mathbf{W}}$} are stable w.r.t. network fine-tuning.} Additionally, these specific frequency components\footref{fn:comp_filter}  also exhibit equivariance to weight scaling and weight permutations.

Therefore, we propose to use the frequency components {\small $\mathcal{F}^{(uv)}_\mathbf{W}$} as the robust watermark. Besides, the overwriting attack is another important issue for watermarking. To defend the watermark from the overwriting attack, we introduce an  additional loss to train the model, which ensures that the overwriting of the watermark will significantly hurt the model's performance.

The contribution of this study can be summarized as follows. (1) We discover and theoretically prove that specific frequency components of a convolutional filter remain invariant during training and are equivariant to weight scaling and weight permutations. (2) Based on the theory, we propose to encode the watermark information to these frequency components, so as to ensure that the watermark is robust to fine-tuning, weight scaling, and weight permutations. (3) Preliminary experiments have demonstrated the effectiveness of the proposed method.

\section{Related Work}
\label{sec:relat}

The robustness of watermarks has always been a key issue in the field of neural network watermarking. In this paper, we limit our discussion to the watermark embedded in network parameters for the protection of the DNN's ownership information. However, fine-tuning, weight scaling, weight permutations, pruning, and distillation may all hurt or remove the watermark from the DNN.

Weight scaling and weight permutations are typical attacking methods, which change the watermark by rearranging the network's parameters. Therefore, \citet{zeng2023huref} found that the multiplication of specific weight matrices were invariant to weight scaling and weight permutations, thereby embedding the watermark information in such multiplication of metrics.  \citet{zhang2024reef} measured the CKA similarity \citep{kornblith2019similarity} between the features of different layers in a DNN as the robust watermark towards weight scaling and weight permutations. 

Compared to the robustness to weight scaling and weight permutations, the robustness to fine-tuning presents a more significant challenge. Up to now, there is no theoretically guaranteed robust watermark to fine-tuning, to the best of our knowledge. Thus, many watermark techniques were implemented in an engineering manner to defend the fine-tuning  attack. \citet{liu2021watermarking} selected network parameters, which did not change a lot during fine-tuning, to encode the watermark information. \citet{tan2023deep} used the classification accuracy on a particular type of adversarial examples, which is termed a trigger set, as the watermark. To enhance robustness, they optimized the trigger set to ensure that the watermarked network could maintain high accuracy on the trigger set, even under the fine-tuning attack. \citet{zeng2023huref} found that the direction of the vector formed by all parameters was relatively stable during fine-tuning, so as to use it to encode watermark information. 

However, \citet{aiken2021neural}, \citet{shafieinejad2021robustness}, and \citet{xu2024instructional} showed that, despite various engineering defense methods, most watermarks could still be effectively removed from the neural network under certain fine-tuning settings. Therefore, a theoretically certificated robust watermark is of considerable value in both theory and practice. To this end, \citet{bansal2022certified} and \citet{ren2023dimension} proposed to use the classification accuracy on a trigger set as the watermark and proved that the classification accuracy was lower bounded when the attacker did not change the network's parameters by more than a distance in terms of {\small $l_{p}$}-norm ({\small $p>1$}). These methods proved a safe range of parameter changes during fine-tuning, but they did not boost the robustness of the watermark or propose an intrinsically robust watermark.

In contrast, we have proved that the convolutional filter's specific frequency components\footref{fn:comp_filter} keep stable during fine-tuning. Thus, we embed the watermark information into these frequency components as a theoretically certificated robust watermark.

\section{Method}
\label{sec:meth}

\subsection{Preliminaries: reformulating the convolution in the frequency domain}
In this subsection, we reformulate the forward propagation through a convolutional filter in the frequency domain. \emph{i.e.}, when we apply a discrete Fourier transform (DFT) to the input feature, and a revised discrete Fourier transform to the convolutional filter, we can get the frequency component vectors at different frequencies for the input feature and the convolutional filter, respectively. As a preliminary, \citet{tang2023defects} have proven that the forward propagation through a convolutional filter can be reformulated as the vector multiplication between the frequency component vectors of the input feature and the convolutional filter at corresponding frequencies.

 Specifically, let us focus on a convolutional filter with {\small$C$} channels and a kernel size of  {\small$K\times K$}. The convolutional filter is parameterized by weights {\small$\mathbf{W}\in\mathbb{R}^{C\times  K\times K  }$} and the bias term {\small$b\in\mathbb{R}$}. Accordingly, we apply this filter to an input feature {\small$\textbf{X}\in\mathbb{R}^{C \times M\times N }$}, and obtain an output feature map {\small$Y\in\mathbb{R}^{M'\times N'}$}.
\begin{equation}\label{eq:layerwise_conv}
\begin{aligned}
   Y =  \mathbf{W}  \otimes  \mathbf{X} + b\cdot \mathbf{1}_{M'\times N'},
\end{aligned}
\end{equation}
where $\otimes$ denotes the convolution operation. {\small$\textbf{1}_{M'\times N'}$} is an {\small$M'\times N'$} matrix, in which elements are all ones. 

\begin{figure}[t]
\vskip 0.2in
\begin{center}
\centerline{\includegraphics[width=\columnwidth]{ 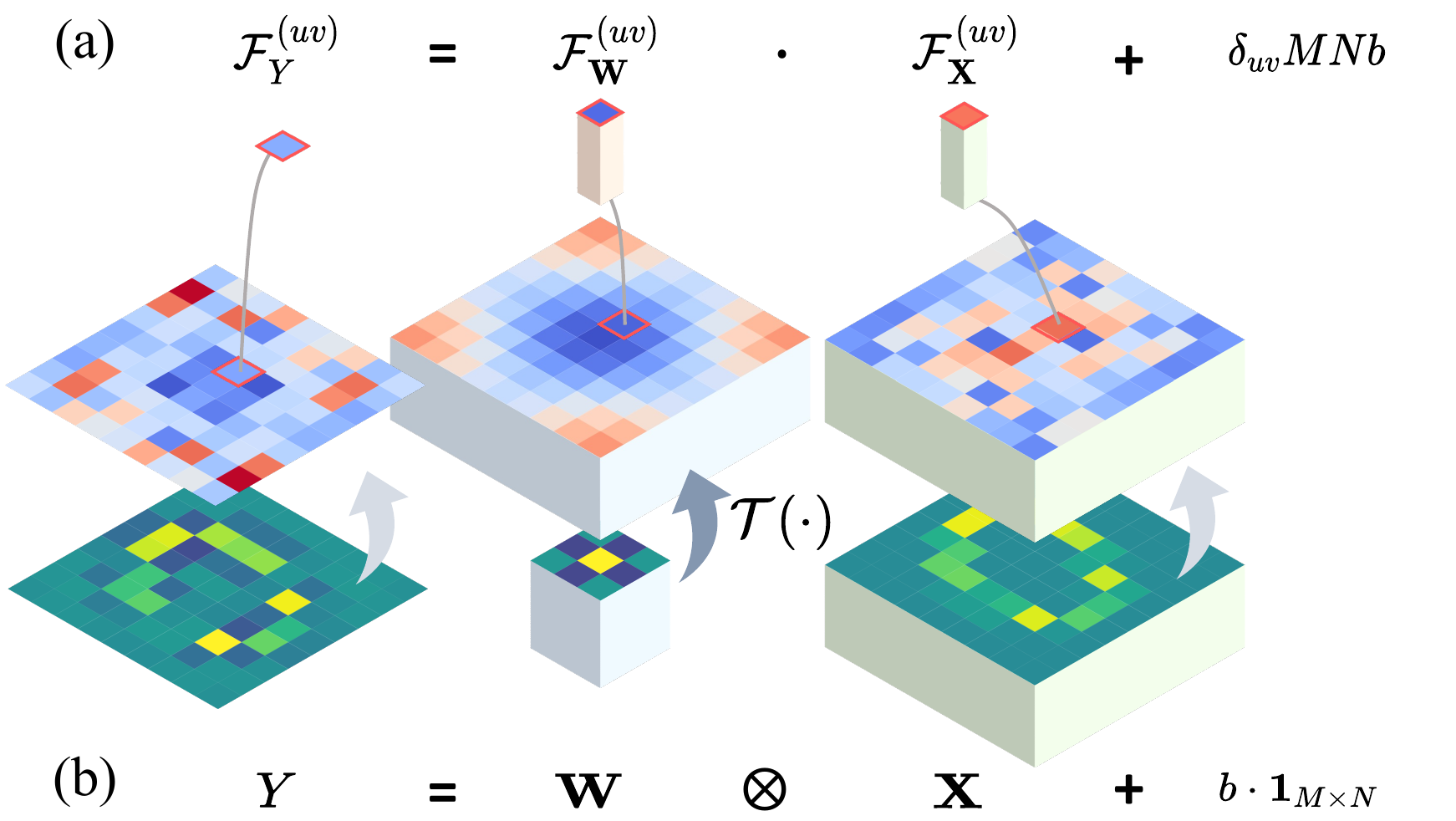}}
\caption{Forward propagation in the frequency domain (a) and forward propagation in the spatial domain (b). The convolution operation with a convolutional filter on the input feature {\small $\mathbf{X}$} is essentially equivalent to a vector multiplication on the frequency components of the input. }
\label{fig_component}
\end{center}
\vskip -0.2in
\end{figure}

\textbf{Frequency components of the input feature and the output feature.} In this way, \citet{tang2023defects} have proven Theorem~\ref{th:forward}, showing that the above forward propagation in Equation~(\ref{eq:layerwise_conv}) can be reformulated as the vector multiplication in the frequency domain as shown in Figure~\ref{fig_component}. However, before that, let us first introduce the notation of frequency components.

Given the input feature {\small$\textbf{X}\in\mathbb{R}^{C \times M\times N }$} and the output feature {\small$Y\in\mathbb{R}^{M\times N}$}, we conduct the two-dimensional DFT on each {\small$c$}-th channel {\small$X^{(c)}\in\mathbb{R}^{M\times N }$} of {\small$\textbf{X}$} and the matrix {\small$Y$} to obtain the frequency element {\small$G_{uv}^{(c)}\in \mathbb{C}$} and  {\small$H_{uv}\in \mathbb{C}$} at frequency {\small$(u,v)$} as follows. {\small $\mathbb{C}$} denotes the set of complex numbers.
\begin{equation}\label{eq:com_feature}
\begin{aligned}
     G_{uv}^{(c)} &=\sum\nolimits_{m=0}^{M-1}\sum\nolimits_{n=0}^{N-1} X_{mn}^{(c)} e^{-i(\frac{um}{M}+\frac{vn}{N})2\pi},\\
    H_{uv} &=\sum\nolimits_{m=0}^{M-1}\sum\nolimits_{n=0}^{N-1} Y_{mn} e^{-i(\frac{um}{M}+\frac{vn}{N})2\pi},
\end{aligned}
\end{equation}
where {\small$X_{mn}^{(c)} \in \mathbb{R}$} denotes the element of {\small$X^{(c)}\in\mathbb{R}^{M\times N}$} at position {\small$(m, n)$}, and {\small$Y_{mn} \in \mathbb{R}$} denotes the element of {\small$Y$}.

For clarity, we can organize all frequency elements belonging to the same {\small $c$}-th channel to construct a frequency spectrum matrix {\small $G^{(c)}\in \mathbb{C}^{M\times N}$}. Alternatively, we can also re-organize these frequency elements at the same frequency {\small $(u,v)$} to form a frequency component vector {\small$\mathcal{F}_{\mathbf{X}}^{(uv)}\in \mathbb{C}^{C} $}. 
\begin{equation}\label{eq:organize}
\begin{aligned}
         \forall c, \quad G^{(c)} &= \begin{bmatrix}
                        G^{(c)}_{00}& \cdots\\
                      \vdots & \ddots 
                    \end{bmatrix} \in \mathbb{C}^{M\times N}, \\
            \forall u,v, \quad \mathcal{F}_{\mathbf{X}}^{(uv)}&=\left [G^{(1)}_{uv},G^{(2)}_{uv},\ldots,G^{(C)}_{uv} \right ]^{\top} \in  \mathbb{C}^{C}.
\end{aligned}
\end{equation}

In this way, we have
\begin{equation}\label{eq:in_tensor}
\mathcal{F}_{\mathbf{X}}= [G^{(1)},G^{(2)},\ldots,G^{(C)}] = \begin{bmatrix}
  \mathcal{F}_{\mathbf{X}}^{(00)}& \cdots \\
  \vdots & \ddots
\end{bmatrix} \in \mathbb{C}^{C\times M \times N}.
\end{equation}
Similarly, {\small$\mathcal{F}_{Y}^{(uv)}=H_{uv}\in \mathbb{C}$} represents the frequency component of the output feature {\small$Y$} at the frequency {\small$(u,v)$}.

For all frequency components {\small $\mathcal{F}_{\mathbf{X}}^{(uv)}$} and  {\small$\mathcal{F}_{Y}^{(uv)}$}, frequency {\small$(u,v)$} close to {\small$(0,0),(0,N-1),(M-1,0)$}, or {\small$(M-1,N-1)$} represents the low frequency, while frequency {\small$(u,v)$} close to {\small$(M/2,N/2)$} is considered as the high frequency.

\begin{theorem}\label{th:forward} (\textbf{Forward propagation in frequency domain})
Based on the above notation,  \cite{tang2023defects} have proven that the forward propagation of the convolution operation in Equation~(\ref{eq:layerwise_conv}) can be reformulated as a vector multiplication in the frequency domain as follows.
\begin{equation}\label{eq:forward_pro}
\begin{aligned}
    Y &=  \mathbf{W}  \otimes  \mathbf{X} + b\cdot \mathbf{1}_{M\times N}  &(\text{Spatial domain}) \\
    &\quad \quad \quad   \Longleftrightarrow \\
    \mathcal{F}_{Y}^{(uv)} &= \mathcal{F}_{\mathbf{W}}^{(uv)} \cdot \mathcal{F}_{\mathbf{X}}^{(uv)} + \delta_{uv}MN b  &(\text{Frequency domain}), \\
\end{aligned}
\end{equation}
where {\small $\cdot$} denotes the scalar product of two vectors;  {\small $\delta_{uv}$} is  defined as {\small $\delta_{uv}=1$}  if and only if {\small$u =  v = 0$}, and {\small $\delta_{uv}=0$} otherwise. In particular, the convolution operation {\small $\otimes$} is conducted with circular padding \citep{jain1989fundamentals} and a stride size of 1, which avoids changing the size of the output feature (\emph{i.e.}, ensuring {\small$M'=M$} and {\small$N'=N$}).
\end{theorem}
\textbf{Frequency components of the convolutional filter.} In Theorem~\ref{th:forward}, {\small$\mathcal{F}_{\mathbf{W}}^{(uv)}$}  represents the frequency component\footref{fn:comp_filter} of the convolutional filter {\small$\mathbf{W}\in \mathbb{R}^{C\times K \times K}$} at frequency {\small$(u,v)$}, and can be obtained by conducting the revised discrete Fourier transform of frequency {\small$(u,v)$}{\small$\mathcal{T}_{uv}(\cdot)$} on {\small$\mathbf{W}$}  as follows.
\begin{equation}\label{eq:com_filter}
    \mathcal{F}_{\mathbf{W}}^{(uv)} = \mathcal{T}_{uv}(\mathbf{W}),
\end{equation}
where {\small$\mathcal{T}_{uv}(\mathbf{W}) = [Q_{uv}^{(1)}, Q_{uv}^{(2)}, \ldots, Q_{uv}^{(C)}]^\top$}, {\small$Q_{uv}^{(c)} = \sum\nolimits_{t=0}^{K-1} \sum\nolimits_{s=0}^{K-1} W_{ts}^{(c)} e^{i(\frac{ut}{M} + \frac{vs}{N}) 2\pi}$}, {\small$ W_{ts}^{(c)}$} denotes the element at position {\small$(t,s)$} of the {\small$c$}-th channel {\small$W^{(c)}\in\mathbb{R}^{K\times K}$} of {\small$\textbf{W}$}.

Similar to Equation~(\ref{eq:in_tensor}), we can define the revised discrete Fourier transform of all frequencies {\small$\mathcal{T}(\cdot)$} by organizing the filter's frequency components to get the frequency tensor as {\small$\mathcal{F}_{\mathbf{W}} = \mathcal{T}(\mathbf{W})$}, where {\small $\mathcal{T}(\mathbf{W})= \begin{bmatrix}
  \mathcal{F}_{\mathbf{W}}^{(00)}& \cdots \\
  \vdots & \ddots
\end{bmatrix} \in \mathbb{C}^{C\times M \times N}$}.

\subsection{Invariant frequency components of the convolutional filter}

In this subsection, we aim to prove that frequency components of the convolutional filter {\small$\mathcal{F}_{\mathbf{W}}^{(uv)}$} at certain frequencies {\small $(u,v)$} are relatively stable during training. Additionally,  these frequency components are also equivariant to other attacks like weight scaling and weight permutations. In this way, we can embed the watermarks into these components to enhance their resistance to fine-tuning, weight scaling, and weight permutations.

\textbf{Specific frequency components of the filter are invariant towards fine-tuning.} Specifically, based on the forward propagation in the frequency domain formulated in Equation~(\ref{eq:forward_pro}), we prove that if the input feature  {\small $\mathbf{X}$} contains only the fundamental frequency components, \emph{i.e.}, {\small $\forall (u,v)\ne (0,0), \mathcal{F}_{\mathbf{X}}^{(uv)} = 0$},  then frequency components {\small $\mathcal{F}_{\mathbf{W}}^{(uv)}$}  at specific frequencies  will not change over the training process.

To prove the invariance of the  frequency components towards fine-tuning, we decompose the entire training process into massive steps of gradient descent optimization. Each step of gradient descent optimization \emph{w.r.t.} the loss function can be formulated as {\small $\mathbf{W}' = \mathbf{W} -\eta \frac{\partial \textit{Loss}}{\partial \mathbf{W}}$}. Let {\small $\mathcal{F}_{\mathbf{W}}^{(uv)} = \mathcal{T}_{uv}(\mathbf{W})$}  and  {\small $\mathcal{F}_{\mathbf{W}'}^{(uv)} = \mathcal{T}_{uv}(\mathbf{W} -\eta \frac{\partial \textit{Loss}}{\partial \mathbf{W}})$} according to Equation~(\ref{eq:com_filter}) denote the frequency components which are extracted  from the filter {\small $\mathbf{W}$} before the step of gradient descent optimization and that extracted from {\small $\mathbf{W}'$} after the optimization, respectively.
\begin{theorem}\label{th:backward_pro}
(\textbf{The change of frequency components during training}, proven in Appendix~\ref{app:sec:backward_pro}) The change of each frequency component {\small $\mathcal{F}_{\mathbf{W}}^{(uv)}$}  before and after a single-step gradient decent optimization is reformulated as follows. 
\begin{equation}\label{eq:backward_pro}
\begin{aligned}
        \Delta \mathcal{F}_{\mathbf{W}}^{(uv)} 
        & = \mathcal{T}_{uv}(\mathbf{W} -\eta \frac{\partial \textit{Loss}}{\partial \mathbf{W}}) - \mathcal{T}_{uv}(\mathbf{W})  \\
        &= \mathcal{F}_{\mathbf{W}'}^{(uv)}-\mathcal{F}_{\mathbf{W}}^{(uv)} \\
        &= - \eta 
        \sum\limits_{u'=0}^{M-1}\sum\limits_{v'=0}^{N-1}A_{uvu'v'} \frac{\partial \textit{Loss}}{\partial \overline{\mathcal{F}}_{Y}^{(u'v')}} \cdot  \overline{\mathcal{F}}_{\mathbf{X}}^{(u'v')},
\end{aligned}
\end{equation}
where  {\small$A_{uvu'v'}=\frac{\sin(\frac{K(u-u')\pi}{M})}{\sin(\frac{(u-u')\pi}{M})} \frac{\sin(\frac{K(v-v')\pi}{N})}{\sin(\frac{(v-v')\pi}{N})} \cdot e^{i(\frac{(K-1)(u-u')}{M}+}$} {$\ ^{\frac{(K-1)(v-v')}{N})\pi} \in\mathbb{C}$} is a complex coefficient; 
 {\small$\overline{\mathcal{F}}_{\mathbf{X}}^{(u'v')}$} denotes the conjugate of {\small$\mathcal{F}_{\mathbf{X}}^{(u'v')}$}.
\end{theorem}

Corollary \ref{co:zero} shows that if  the input feature  {\small $\mathbf{X}$} only contains the fundamental frequency component, then the specific frequency components {\small $\mathcal{F}_{\mathbf{W}}^{(uv)}$} keep unchanged over the training process.

\begin{corollary}\label{co:zero} (\textbf{Invariant frequency components towards fine-tuning}, proven in Appendix~\ref{app:sec:zero}) In the training process, if the input feature {\small $\mathbf{X}$} only contains the fundamental frequency component, \emph{i.e.}, {\small $\forall (u,v) \ne (0,0),\mathcal{F}_{\mathbf{X}}^{(uv)}=0$}, then frequency components {\small $\mathcal{F}_{\mathbf{W}}^{(uv)}$} at the following frequencies in the set {\small $S$} keep invariant.
\begin{equation}\label{eq:zero}
    \forall (u, v) \in S, \quad \Delta \mathcal{F}_{\mathbf{W}}^{(uv)} = \mathbf{0}, 
\end{equation}
where {\small $S$} is the set of the  specific frequencies, defined as {\small $S = \{ (u, v) \ | \ u = iM/K \ \text{or} \ v = jN/K; \ i, j \in \{1, 2, \dots, K-1\} \}$}.
\end{corollary}

Corollary~\ref{co:zero} indicates an ideal case where the input feature only contains the fundamental frequency component, and {\small $u, v$} can take non-integer values. In this case, {\small $\Delta \mathcal{F}_{\mathbf{W}}^{(uv)}$} at frequencies in the set {\small $S$} are strictly zero vectors.

However, in real applications, the input feature usually contains low-frequency components, and frequencies must be integers when conducting the DFT. Under these conditions, each element in {\small $\Delta \mathcal{F}_{\mathbf{W}}^{(uv)}$} is nearly zero at integer frequencies that are close to the frequencies in the set {\small $S$}.

\begin{proposition}\label{pro:zero}
In the training process, if the input feature {\small $\mathbf{X}$} only contains the low-frequency components, \emph{i.e.}, {\small $\forall (u,v) \notin S^{\text{low}}_{r},  \mathcal{F}_{\mathbf{X}}^{(uv)}=0$},  where {\small $S^{\text{low}}_{r}  =  \{(u,v) | u\in [0,r] \cup [M-r, M),  v\in [0,r] \cup [N-r, N) \} $} and  {\small $r$} is a positive integer {\small $(r\le 2)$}, then frequency components {\small $\mathcal{F}_{\mathbf{W}}^{(uv)}$} at the following frequencies in the set {\small $S'$} keep relative stable.
\begin{equation}\label{eq:nearly_zero}
    \forall (u, v) \in S', \quad \Delta \mathcal{F}_{\mathbf{W}}^{(uv)} \approx  \mathbf{0}, 
\end{equation}
where {\small $S'$} is the set of the specific integer frequencies, defined as {\small $S' = \{ (u, v) |  u = \lfloor iM/K \rceil  \ \text{or} \ v = \lfloor jN/K \rceil;\  i, j \in \{1, 2, \dots, K-1\} \}$}, {\small $\lfloor x \rceil$} is  used to round the real number {\small $x$} to the nearest integer.
\end{proposition}

\textbf{Towards weight scaling.} Weight scaling attack means scaling the weights of a  convolutional layer by a constant {\small $a$}, and scaling the weights of the next convolutional layer by the inverse proportion {\small $1/a$}. In this way, the model's performance will not be affected, but the watermark embedded in the weights usually will change. Theorem~\ref{th:scale} shows that the frequency components are equivariant to the weight scaling attack. 

\begin{theorem}\label{th:scale} (\textbf{Equivariance towards weight scaling}, proven in Appendix~\ref{app:sec:scale}) If we scale all weights in the convolutional filter {\small $\mathbf{W}$} by a constant {\small $a$} as {\small $\mathbf{W}^* = a \cdot \mathbf{W}(a>0)$}, then the frequency components of {\small $\mathbf{W}^*$} are equal to the scaled frequency components of {\small $\mathbf{W}$}, as follows.
\begin{equation}\label{eq:scale}
     \forall a,u,v,\quad \mathcal{F}^{(uv)}_{\mathbf{W}^*} = a \cdot \mathcal{F}^{(uv)}_{\mathbf{W}}.
\end{equation}
\end{theorem}

\textbf{Towards weight permutations.} Permutation attack on the filters means permuting the filters and corresponding bias terms of a convolutional layer, and then permuting the channels of every filter of the next convolutional layer in the same order. As a result, the network’s outputs remain unaffected, while the watermark embedded in the weights is usually altered.

We further investigate the equivariance of the frequency components when permuting the convolutional filters. Given a convolutional layer with {\small $D$} convolutional filters with {\small $D$} bis terms arranged as  {\small $\mathbb{W} =  [\mathbf{W}_{ 1}, \mathbf{W}_{ 2},\cdots,   \mathbf{W}_{ D} ]$} and {\small $\mathbf{b} =  [b_{ 1}, b_{2},\cdots,   b_{D} ]$}.

\begin{theorem}\label{th:permu}
(\textbf{Equivariance towards weight permutations}, proven in Appendix~\ref{app:sec:permu}) If we use a permutation {\small $\pi$} to rearrange the above filters and bias terms as  {\small $\pi \mathbb{W}  =  [\mathbf{W}_{ \pi(1)}, \mathbf{W}_{ \pi(2)},\cdots,   \mathbf{W}_{ \pi(D)} ]$} and {\small $\pi\mathbf{b} =  [b_{ \pi(1)}, b_{\pi(2)},\cdots,   b_{\pi(D)} ]$}, where  {\small $[ \pi(1),\pi(2),\cdots,\pi(D) ]$} is a random permutation of integers from {\small $1$} to {\small $D$}, then the frequency components of {\small $\pi \mathbb{W}$} are equal to the permuted frequency components of {\small $\mathbb{W}$}, as follows.
\begin{equation}
     \forall \pi,u,v, \ \left [ \mathcal{F}^{(uv)}_{\mathbf{W}_{\pi(1)}},\cdots, \mathcal{F}^{(uv)}_{\mathbf{W}_{\pi(D)}} \right ] = \pi
     \left [\mathcal{F}_{ \mathbf{W}_{1}}^{(uv)} \cdots \mathcal{F}_{ \mathbf{W}_{D}}^{(uv)}   \right ],
\end{equation}
where {\small $\mathcal{F}_{\mathbf{W}_{ d}}^{(uv)} = \mathcal{T}_{uv}(\mathbf{W}_{d}) \in \mathbb{C}^{C}$} denote the frequency components extracted from the {\small $d$}-th filter {\small $\mathbf{W}_{d}$} at frequency {\small $(u,v)$}. 
\end{theorem}

\begin{figure}[t]
\vskip 0.2in
\begin{center}
\centerline{\includegraphics[width=\columnwidth]{ 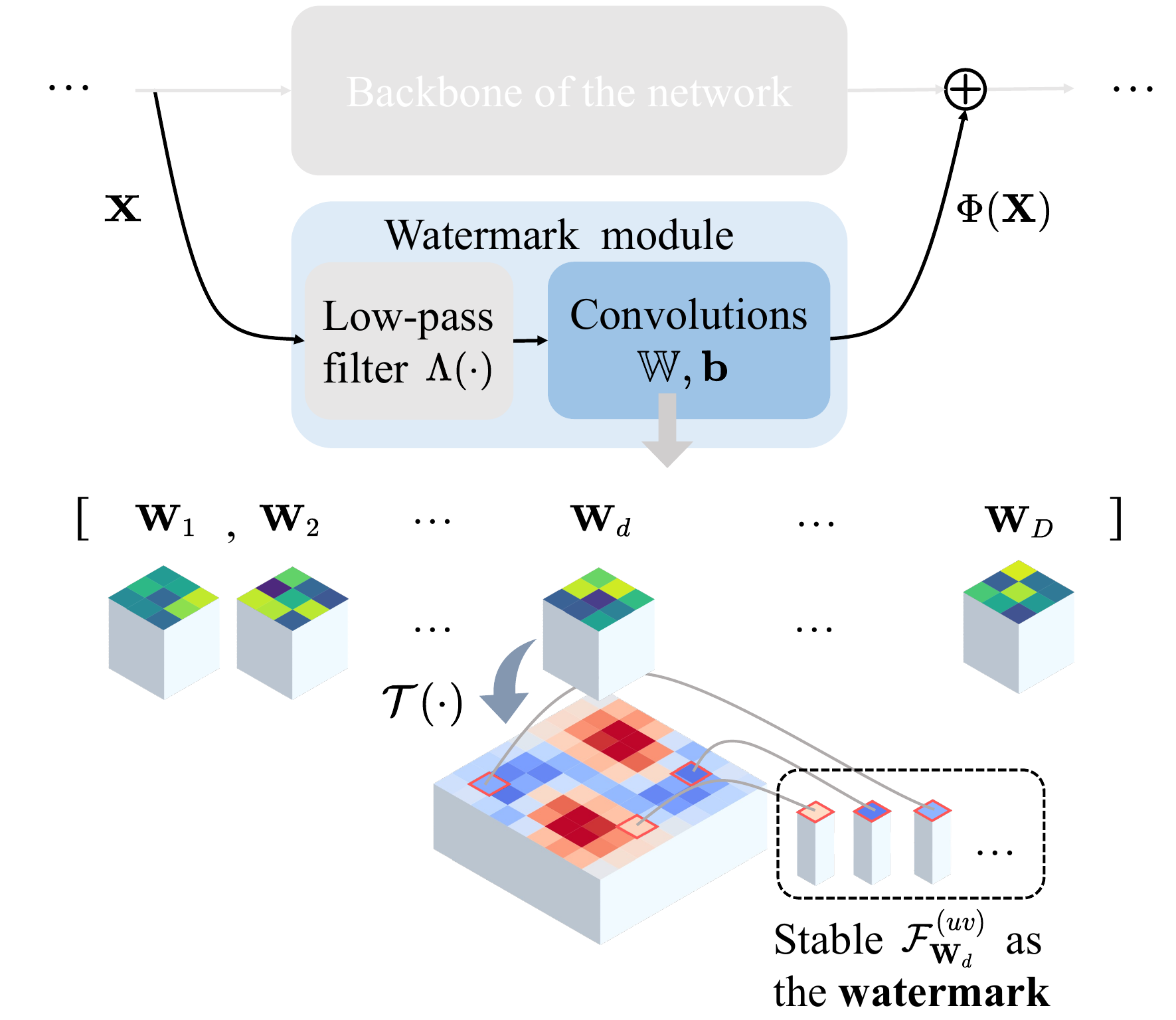}}
\caption{The architecture of the watermark module.  The watermark module is connected in parallel to the backbone of the neural network. We extract the specific frequency components from the convolutional filters in the watermark module as the network's watermark. }
\label{fig_module}
\end{center}
\vskip -0.2in
\end{figure}

\subsection{Using the invariant frequency components as the neural network's watermark}

In the last subsection, we prove that if the input feature only contains the  low-frequency components, the filter's frequency components {\small $\mathcal{F}^{(uv)}_{\mathbf{W}}$} at specific frequencies {\small $(u, v)$} keep stable during training. Furthermore, these components exhibit equivariance to weight scaling and weight permutations.

\textbf{Watermark module.} All the above findings and proofs enable us to use the specific frequency components as the watermark of the neural network. In this way, the watermark will be highly robust to fine-tuning, weight scaling, and weight permutations.  Specifically, as Figure~\ref{fig_module} shows, we construct the following watermark module {\small $\Phi(\mathbf{X})$} to contain the watermark, which consists of a low-pass filter {\small $\Lambda(\cdot)$} and convolution operations (with {\small $D$} convolutional filters {\small $\mathbb{W} = [\mathbf{W}_{ 1}, \mathbf{W}_{ 2},\cdots,   \mathbf{W}_{ D} ]$} and {\small $D$} bias terms {\small $\mathbf{b} =  [b_{ 1}, b_{2},\cdots,   b_{D}]$}). 
\begin{equation}\label{eq:module}
\begin{aligned}
    \Phi(\mathbf{X}) &= \left [Y_{1}, Y_{2},\cdots,   Y_{D}\right ], \\
  \emph{\text{s.t.}}\quad  Y_{d} &= \mathbf{W}_{d} \otimes \Lambda(\mathbf{X}) + b_{d} \cdot \mathbf{1}_{M\times N},
\end{aligned}
\end{equation}
where the low-pass filtering operation {\small $\Lambda(\cdot)$} preserves frequency components in {\small $\mathbf{X}$} at low frequencies in {\small$ S^{\text{low}}_{r} = \{(u,v) | u\in [0,r] \cup [M-r, M),  v\in [0,r] \cup [N-r, N) \}$} {\small$(r\le 2)$}  and removes all other frequency components, \emph{i.e.}, setting {\small $\forall (u,v) \notin S^{\text{low}}_{r}, \mathcal{F}_{\mathbf{X}}^{(uv)}=0$}. {\small $\mathbf{1}_{M\times N}$} is an {\small $M\times N$} matrix, in which elements are all ones.

\textbf{Invariant frequency components as the watermark. } \textit{In this way, when we extract frequency components {\small $\mathcal{F}_{\mathbf{W}_{d}}^{(uv)}$} from every {\small $d$}-th convolutional filter {\small $\mathbf{W}_{d}$} in the watermark module  {\small $\Phi(\mathbf{X})$} based on Equation~(\ref{eq:com_filter}), we can consider the  frequency components at the following frequencies in the set {\small $S'$}, as the watermark.}
\begin{equation}\label{eq:watermark}
    \left [ \mathcal{F}_{\mathbf{W}_{1}}^{(uv)}, \mathcal{F}_{\mathbf{W}_{2}}^{(uv)},\cdots,   \mathcal{F}_{\mathbf{W}_{D}}^{(uv)} \right ]\quad   s.t.\quad  (u,v) \in S', 
\end{equation}
\textit{where {\small $S' = \{ (u, v) |  u = \lfloor iM/K \rceil  \ \text{or} \ v = \lfloor jN/K \rceil;\  i, j \in \{1, 2, \dots, K-1\} \}$}. According to Proposition~\ref{pro:zero}, the watermark will keep stable during training. }

\textit{Implementation details.} We notice that in the watermark module, the low-pass filter {\small $\Lambda(\cdot)$} may hurt  the flexibility of feature representations. Therefore, as Figure~\ref{fig_module} shows, the watermark module is connected in parallel to the backbone architecture of the neural network. In this way, this design does not significantly change the network's architecture or seriously hurt its performance. In this paper, unless stated otherwise, we set the integer {\small $r = 1$}, the kernel size {\small $K = 3$}.

\begin{figure}[t]
\vskip 0.2in
\begin{center}
\centerline{\includegraphics[width=\columnwidth]{ 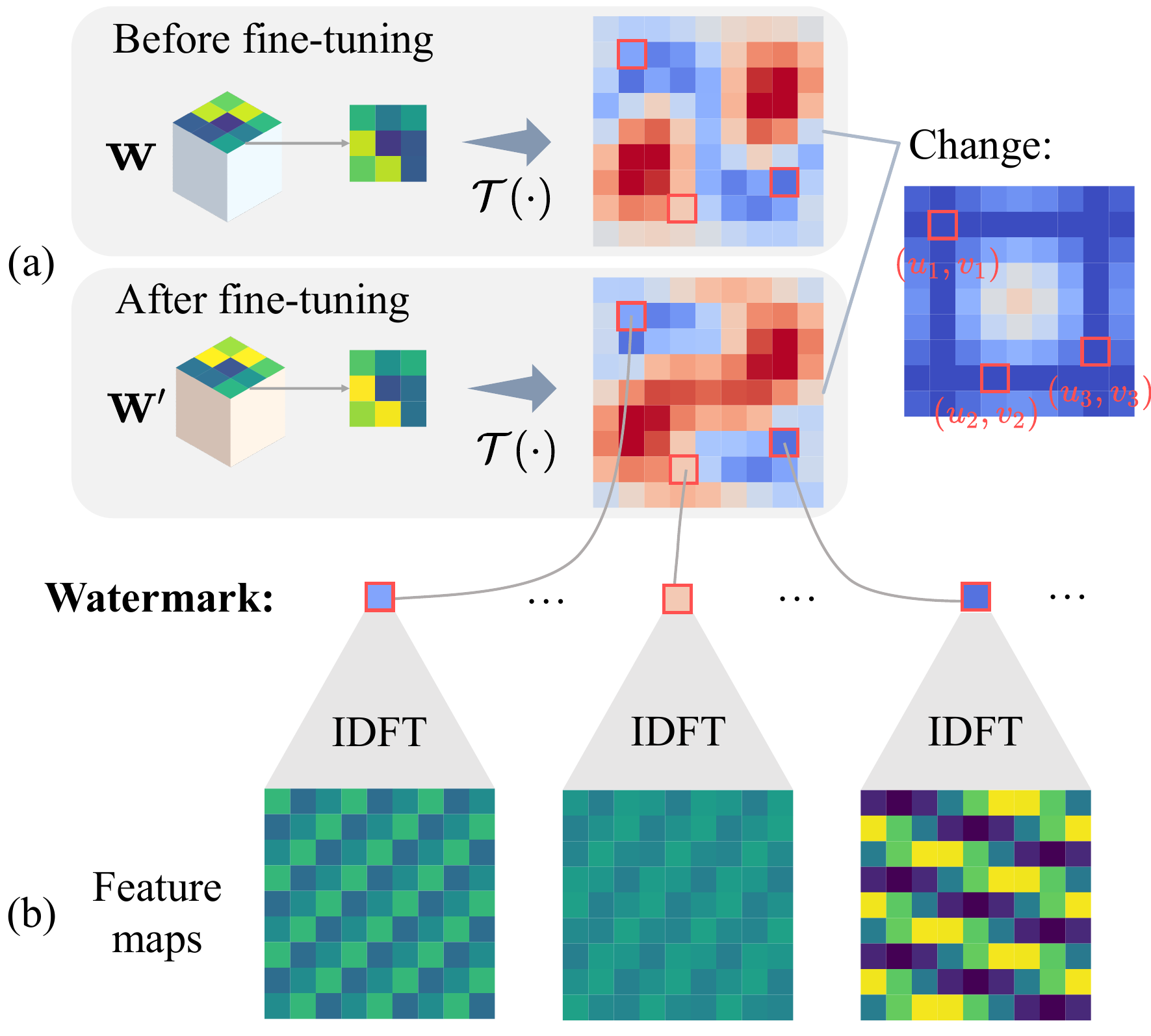}}
\caption{Visualization of the watermark. (a) shows the specific frequencies in the set {\small $S'$} used as the watermark. (b) shows the feature maps when we apply the inverse discrete Fourier transform (IDFT) to some unit frequency components used as the watermark. The frequency components are extracted from a single channel of a 
 {\small $3 \times 3$} convolutional filter in the watermark module and the input feature map has a width and height of {\small $9 \times 9$}, so the set {\small $S' = \{ (u, v) |  u =  3i\ \text{or} \ v = 3j;\  i, j \in \{1, 2\} \}$}. For clarity, we move low frequencies to the center of the spectrum map, and move high frequencies to corners of the spectrum map.
}
\label{fig_watermark}
\end{center}
\vskip -0.2in
\end{figure}

\textbf{Visualization of the watermark.} Figure~\ref{fig_watermark}(a) shows the specific frequencies in the set {\small $S'$} used as the watermark. Figure~\ref{fig_watermark}(b) shows the feature maps when we apply the inverse discrete Fourier transform (IDFT) to some unit frequency components which are used as the watermark. Since our transform of the convolutional filter in Equation~(\ref{eq:com_filter}) is irreversible, we use the feature maps here only to illustrate the characteristics of the frequency components\footref{fn:comp_filter} in the spatial domain.

\subsection{Detecting the watermark} \label{subsec:detect}
In this subsection, we introduce how to detect the watermark. Given a source watermarked DNN with a watermark module containing {\small $D$} convolutional filters {\small $[\mathbf{W}_{ 1}, \mathbf{W}_{ 2},\cdots,   \mathbf{W}_{ D} ]$} and a suspicious DNN with a watermark module containing {\small $D$} convolutional filters {\small $[\mathbf{W}'_{1}, \mathbf{W}',\cdots,   \mathbf{W}'_{ D} ]$}, we aim to detect  whether the suspicious DNN is obtained from the source DNN by fine-tuning, weight scaling or weight permutations.

Considering the permutation attack, the detection towards the  frequency components should consider  the matching between the frequency components of different  convolutional filters of the two DNNs, \emph{i.e.}, we can definitely  find a permutation {\small $[ \pi(1),\pi(2),\cdots,\pi(D) ]$} to assign each  {\small $d$}-th convolutional filter  {\small $\mathbf{W}_d$} in the source DNN with the {\small $\pi(d)$}-th filter {\small $\mathbf{W}'_{\pi(d)}$} in the suspicious DNN. Specifically, we use the following watermark detection rate {\small $\text{DR}$} between two DNNs to identify the matching quality.
\begin{equation}\label{eq:detect}
    \text{DR} = \frac{\sum_{(u,v)\in S',d} \mathbb{I}(\text{cos}(\mathcal{F}^{uv}_{\mathbf{W}_d}, \mathcal{F}^{uv}_{\mathbf{W}'_{\pi(d)}}) \ge \tau)}{Z}\times 100\%
\end{equation}
where  {\small $\mathbb{I}$} denotes an indicator function that equals {\small $1$} if {\small $\text{cos}(\mathcal{F}^{(uv)}_{\mathbf{W}_d}, \mathcal{F}^{uv}_{\mathbf{W}'_{\pi(d)}})\ge \tau$} and {\small $0$} otherwise.  {\small $\text{cos}(\mathcal{F}^{(uv)}_{\mathbf{W}_d}, \mathcal{F}^{uv}_{\mathbf{W}'_{\pi(d)}})$}  denotes the cosine similarity \footnote{\label{fn:cos} The cosine similarity {\small $\text{cos}(\mathbf{z}_1,\mathbf{z}_2)$} between two complex vectors {\small $\mathbf{z}_1$} and {\small $\mathbf{z}_2$} is defined as {\small $\frac{\text{Re}(\overline{\mathbf{z}_1} \cdot \mathbf{z}_2)}{\|\mathbf{z}_1\| \|\mathbf{z}_2\|}$}, where {\small $\overline{\mathbf{z}_1}$} denotes the conjugate of {\small $\mathbf{z}_1$}, {\small $\|\mathbf{z}_1\|$} denotes the magnitude of {\small $\mathbf{z}_1$}, and {\small $\text{Re}(\cdot)$} represents the real part of a complex number. The cosine similarity, which ranges from {\small $[-1,1]$}, measures the directional similarity between two complex vectors. When {\small $\text{cos}(\mathbf{z}_1,\mathbf{z}_2) = 1$}, {\small $\mathbf{z}_1$} and {\small $\mathbf{z}_2$} have the same direction, while when {\small $\text{cos}(\mathbf{z}_1,\mathbf{z}_2) = -1$}, {\small $\mathbf{z}_1$} and {\small $\mathbf{z}_2$} have opposite directions.} of the frequency components. {\small$Z$} denotes the total number of the frequency components that are used  as the watermark. {\small $\tau$} is a threshold, and we set {\small $\tau=0.995$} in this paper unless otherwise stated. If the watermark detection rate {\small $\text{DR}$} is high, we can determine that the suspicious network originates from the source network.

\subsection{Learning towards the overwriting attack} \label{subsec:overwriting}
Another challenge is the resistance of the watermark to the overwriting attack. Given the watermark module's architecture, if the attacker has obtained the authority to edit its parameters, then he can overwrite the weights $\mathbb{W}$ in the watermark module with entirely new values, so as to change the watermark.

Let us consider a DNN for the classification of {\small $n$} categories. To defend the overwriting attack, the basic idea is to construct the {\small $(n+1)$}-th category as a pseudo category besides the existing {\small $n$} categories.  If the neural network is not attacked, it is supposed to classify input samples normally. Otherwise, if the network is under an overwriting attack, then it is supposed to classify all samples into the pseudo category. In this way, overwriting the watermark  will significantly hurt the classification performance of the DNN.  Therefore, we train the network by adding an additional loss {\small $\mathcal{L}_{\text{attack}}$}, which pushes the attacked network to classify all samples into the pseudo category,  to the standard cross-entropy loss {\small $\mathcal{L}_{\text{CE}}$} for multi-category classification.
\begin{equation}\label{eq:loss}
\begin{aligned}
        \mathcal{L}(\mathbb{W}, \mathbf{b}, \theta|x)\  =&\ \  \mathcal{L}_{\text{CE}}(\mathbb{W}, \mathbf{b}, \theta|x) + \mathcal{L}_{\text{attack}}(\mathbb{W}, \mathbf{b}, \theta|x) \\
        = &- \sum_{k=1}^{n} p(y=k|x) \text{log}\ q(y=k|x;\mathbb{W}, b, \theta) \\
        &- \lambda \cdot \text{log}\ q(y=n+1|x;\mathbb{W}+\epsilon, b, \theta), 
\end{aligned}
\end{equation}
where {\small $x$} denotes an input sample, and {\small $y$} denotes its corresponding label; {\small $\theta$} denotes the network's parameters. {\small $q(y=k|x;\mathbb{W}, \mathbf{b}, \theta)$} denotes the classification probability predicted by the neural network. {\small $p(y=k|x)$} is the ground truth probability. The scalar weight {\small $\lambda$} balances the influence of {\small $\mathcal{L}_{\text{CE}}$} and {\small $\mathcal{L}_{\text{attack}}$}.

In the above loss function, we add a random noise\footnote{The magnitude and other specific settings of the noise {\small $\epsilon$} will be introduced later.} {\small $\epsilon$} to the parameters {\small $\mathbb{W}$} in the watermark module to mimic the state of the neural network with overwritten parameters. To enhance the module's sensitivity to such attacks, we do not completely overwrite the parameters but add random noise.

\textit{Ablation studies.} We conducted an ablation experiment to evaluate the effectiveness of the newly added loss term {\small $\mathcal{L}_{\text{attack}}$}, \emph{i.e.}, examining whether the performance of the neural network was significantly hurt under the overwriting attack when the network was trained with the loss function {\small $\mathcal{L}$} in Equation~(\ref{eq:loss}). We compared the classification accuracy of the network without the attack and the classification accuracy  under the attack to analyze the performance decline of the network towards the overwriting attack.

We ran experiments of AlexNet \citep{krizhevsky2012imagenet} and ResNet18 \citep{he2016deep} on Caltech-101, Caltech-256 \citep{fei2006one},  CIFAR-10 and CIFAR-100 \citep{krizhevsky2009learning} for image classification tasks. For AlexNet, the watermark module containing {\small $256$} convolutional filters was connected to the third convolutional layer. For ResNet18, the watermark module containing {\small $256$} convolutional filters was connected to the second convolutional layer of the second residual block. The scalar weight {\small $\lambda$} was set to {\small $5\times 10^{-4}$}. The noise {\small $\epsilon$} added to the parameters in the watermark module was obtained by conducting the IDFT on a unit frequency component at a random frequency, and the {\small $l_2$}-norm of the noise {\small $\epsilon$} was set to {\small $0.5$} times the {\small $l_2$}-norm of the weights. 

Table~\ref{exp:overwriting} shows the experiment results. We observe that if the network is trained with the loss function {\small $\mathcal{L}=\mathcal{L}_{CE} + \mathcal{L}_{attack} $} in Equation~(\ref{eq:loss}),  the classification accuracy significantly drops under the overwriting attack. The results indicate that the newly introduced loss term effectively defends the overwriting attack.

\begin{figure}[t]
\vskip 0.2in
\begin{center}
\centerline{\includegraphics[width=\columnwidth]{ 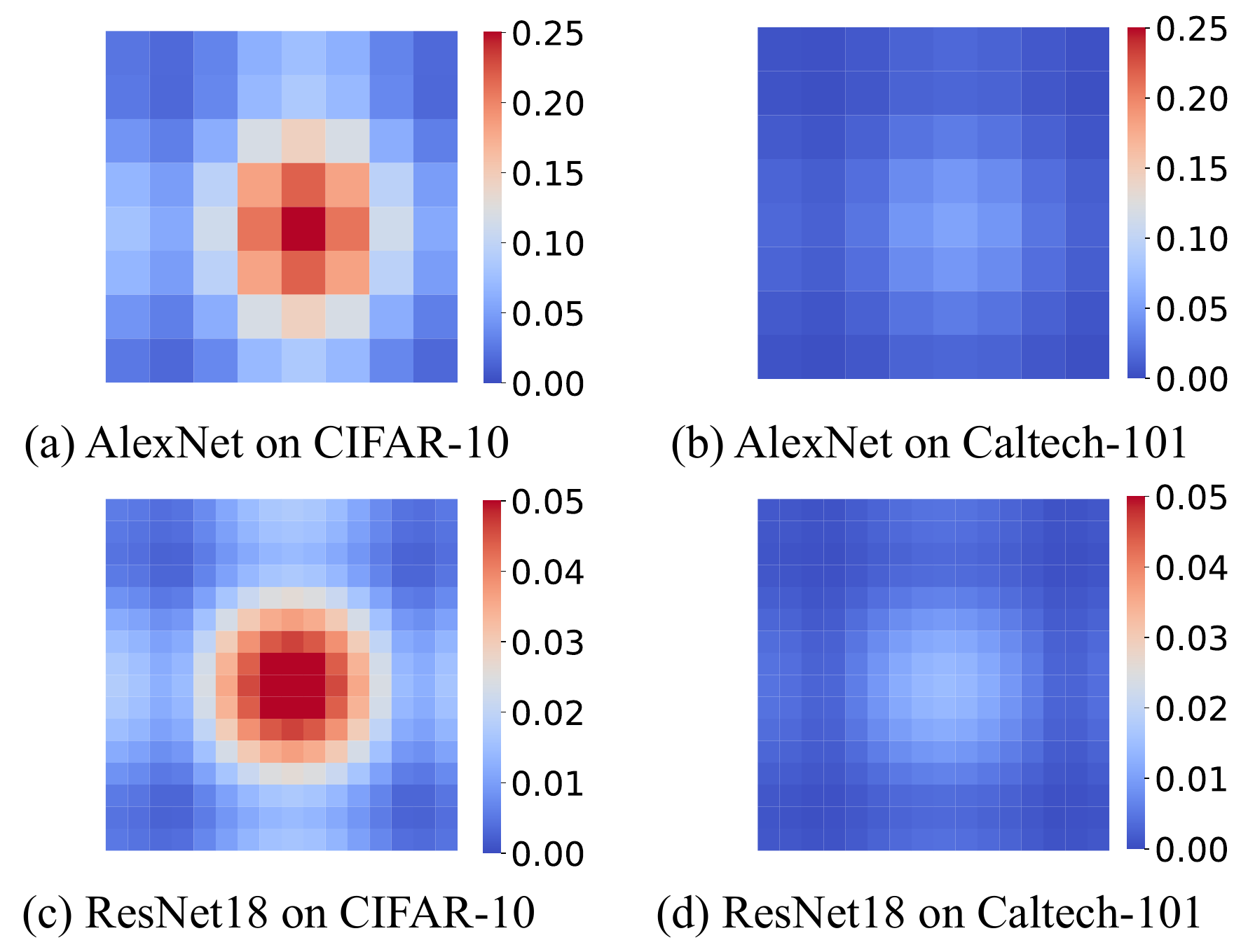}}
\caption{Heatmaps showing the average norm of the change of the frequency components {\small $\mathbb{E}_{d}[ \| \Delta \mathcal{F}_{\mathbf{W}_{d}}^{(uv)} \|]$} before and after fine-tuning at different frequencies {\small $(u,v)$} over all convolutional filters.  For clarity, we move low frequencies to the center of the spectrum map, and move high frequencies to corners of the spectrum map}
\label{fig_fine-tune}
\end{center}
\vskip -0.2in
\end{figure}

\begin{table*}[ht]
\centering
\caption{Experiment results of the effectiveness of the newly added loss term {\small $\mathcal{L}_{attack}$}. Baseline denotes the test accuracy of a neural network normally trained without the watermark. With {\small $\mathcal{L}_{CE} + \mathcal{L}_{attack}$} denotes the test accuracy of a watermarked network trained with the loss function {\small $\mathcal{L}_{CE} + \mathcal{L}_{attack}$}, and With {\small $\mathcal{L}_{CE}$} denotes the test accuracy of a watermarked network trained without the added loss term {\small $\mathcal{L}_{attack}$}. The accuracy outside the bracket represents the accuracy of the network without the overwriting attack, and the accuracy inside the bracket represents the accuracy of the network under the overwriting attack.}
\resizebox{0.7\linewidth}{!}{ 
\begin{tabular}{ccccccc}
\toprule
 \multirow{2}{*}{Dataset} & \multicolumn{2}{c}{Baseline (\%)} & \multicolumn{2}{c}{With $\mathcal{L}_{CE} + \mathcal{L}_{attack}$ (\%)} & \multicolumn{2}{c}{With $\mathcal{L}_{CE}$ (\%)} \\
 \cmidrule(lr){2-3} \cmidrule(lr){4-5} \cmidrule(lr){6-7}
& AlexNet & ResNet-18 & AlexNet & ResNet-18 & AlexNet & ResNet-18 \\
\midrule
 CIFAR-10  & $91.03$ & $94.83$ & $90.28\ (43.55)$ & $92.17\ (72.26)$ & $91.12\ (91.12)$ & $94.89\ (94.89)$ \\
 CIFAR-100 & $68.10$ & $76.29$ & $66.34\ (36.93)$ & $75.49\ (41.18)$ & $67.52\ (67.52)$ & $76.53\ (76.53)$ \\
Caltech-101 & $66.46$ & $70.10$ & $62.15\ (32.53)$ & $67.14\ (41.33)$ & $67.84\ (67.84)$ & $69.87\ (69.87)$ \\
 Caltech-256 & $40.50$ & $54.61$ & $37.97\ (15.37)$ & $50.33\ (18.13)$ & $39.22\ (39.22)$ & $53.86\ (53.86)$ \\
\bottomrule
\label{exp:overwriting}
\end{tabular}}
\end{table*}

\begin{table}[ht]
\centering
\caption{Experiment results of verifying the robustness towards fine-tuning. Baseline denotes the accuracy of a neural network normally trained without the watermark. Ours denotes the accuracy of a watermarked network. The rate inside the bracket denotes the watermark detection rate of the fine-tuned DNN. Accuracy outside the bracket denotes test accuracy on the dataset. }
\resizebox{\columnwidth}{!}{ 
\begin{tabular}{cccccc}
\toprule
\multirow{2}{*}{Source} & \multirow{2}{*}{Target} & \multicolumn{2}{c}{Baseline (\%)} & \multicolumn{2}{c}{Ours (\%)} \\
 \cmidrule(lr){3-4} \cmidrule(lr){5-6} 
 &  & AlexNet & ResNet-18 & AlexNet & ResNet-18  \\
\midrule
\multirow{2}{*}{CIFAR-100} 
& CIFAR-10 & $88.90$  & $93.03$  & $89.65\ (100)$  & $94.12\ (100)$  \\
& Caltech-101 & $70.02$  & $76.80$  & $72.11\ (100)$  & $79.98\ (100)$   \\
\bottomrule
\label{exp:finetune_acc}
\end{tabular}}
\end{table}

\begin{table}[ht]
\centering
\caption{Experiment results of verifying the robustness towards weight scaling. The rate outside the bracket denotes the watermark detection rate without the weight scaling attack, and the rate inside the bracket denotes the watermark detection rate under the weight scaling attack.} 
\resizebox{\columnwidth}{!}{ 
\begin{tabular}{ccccc}
\toprule
$a$ & CIFAR-10 (\%) & CIFAR-100 (\%) & Caltech-101 (\%) & Caltech-256 (\%) \\
\midrule
$10$ & $100\ (100)$  & $100\ (100)$  & $100\ (100)$  & $100\ (100)$   \\
$100$ & $100\ (100)$  & $100\ (100)$  & $100\ (100)$  & $100\ (100)$   \\
\bottomrule
\label{exp:scale}
\end{tabular}}
\end{table}

\begin{table}[ht]
\centering
\caption{Experiment results of verifying the robustness towards weight permutations. The rate outside the bracket denotes the watermark detection rate without the weight permutation attack, and the rate inside the bracket denotes the watermark detection rate under the weight permutation attack.}
\resizebox{\columnwidth}{!}{ 
\begin{tabular}{ccccc}
\toprule
$\pi$ & CIFAR-10 (\%) & CIFAR-100 (\%) & Caltech-101 (\%) & Caltech-256 (\%) \\
\midrule
$\pi_1$ & $100\ (100)$  & $100\ (100)$  & $100\ (100)$  & $100\ (100)$   \\
$\pi_2$ & $100\ (100)$  & $100\ (100)$  & $100\ (100)$  & $100\ (100)$   \\
\bottomrule
\label{exp:permutation}
\end{tabular}}
\end{table}

\subsection{Verifying the robustness of the watermark}

\textbf{Verifying the robustness towards fine-tuning.} 
We conducted the experiments to verify the invariance of the proposed watermark towards fine-tuning. Let us fine-tune a trained DNN with watermark module containing filters {\small $[\mathbf{W}_{ 1}, \mathbf{W}_{ 2},\cdots,   \mathbf{W}_{ D} ]$}, and obtain a fine-tuned DNN with filters {\small $[\mathbf{W}'_{ 1}, \mathbf{W}'_{ 2},\cdots,   \mathbf{W}'_{ D} ]$}. We computed the average the norm of the change of the frequency components {\small $\mathbb{E}_{d}[ \| \Delta \mathcal{F}_{\mathbf{W}_{d}}^{(uv)} \|]$ } to measure the invariance of the proposed watermark, where {\small  $\Delta \mathcal{F}_{\mathbf{W}_{d}}^{(uv)} = \mathcal{F}_{\mathbf{W}'_{d}}^{(uv)}- \mathcal{F}_{\mathbf{W}_{d}}^{(uv)}$ } denoted the change of the frequency components extracted from the {\small $d$}-th convolutional filter.

We trained AlexNet and ResNet18 on CIFAR-100, and then fine-tuned them on CIFAR-10 and Caltech-101. All other experiment settings remained the same as described in Section~\ref{subsec:overwriting}. Figure~\ref{fig_fine-tune} shows the frequency components used as the watermark keep stable during fine-tuning. Table~\ref{exp:finetune_acc} shows that adding a watermark does not decline the fine-tuning performance of the network. The results indicate that the watermark is robust to the fine-tuning attack.

\textbf{Verifying the robustness towards weight scaling.} We conducted experiments to verify the robustness of the proposed watermark towards weight scaling. Given a watermarked DNN, we scaled the parameters in the watermark module by a constant {\small $a (a>0)$}, and then detected the watermark using the method introduced in Section~\ref{subsec:detect}. We used the watermark detection rate {\small $\text{DR}$} to show the robustness of the watermark towards weight scaling. We trained AlexNet on CIFAR-10, CIFAR-100, Caltech-101 and Caltech-256. All other experiment settings remained the same as described in Section~\ref{subsec:overwriting}. Table~\ref{exp:scale} shows the experiment results. All the watermark detection rates are {\small $100\%$}, showing that our method is highly robust to the weight scaling attack. 

\textbf{Verifying the robustness towards weight permutations.} We conducted experiments to verify the robustness of the proposed watermark towards weight permutations. Given a watermarked DNN, we permuted the filters in the watermark module with a  random  permutation {\small $\pi$},  and then detected the watermark using the method introduced in Section~\ref{subsec:detect}. We used the watermark detection rate {\small $\text{DR}$} to show the robustness of the watermark towards weight scaling. We trained AlexNet on CIFAR-10, CIFAR-100, Caltech-101 and Caltech-256. All other experiment settings remained the same as described in Section~\ref{subsec:overwriting}. Table~\ref{exp:scale} shows the experiment results. All the watermark detection rates are {\small $100\%$}, showing that our method is highly robust to the weight permutation attack.

\section{Conclusion}
In this paper, we discover and theoretically prove that specific frequency components of a convolutional filter keep stable during training and have equivariance towards weight scaling and weight permutations. Based on the theory, we propose to use these frequency components as the network's watermark to embed the ownership information. Thus, our proposed watermark technique is theoretically guaranteed to be robust to fine-tuning, weight scaling, and weight permutations. Additionally, to defend against the overwriting attack, we add an additional loss term during training to make sure that the network's performance will drop significantly under the overwriting attack. Preliminary experiments have demonstrated the effectiveness of the proposed method.

\section{Impact Statements}
This paper presents work whose goal is to advance the field of Machine Learning. There are many potential societal consequences of our work, none which we feel must be specifically highlighted here.

\bibliographystyle{plainnat}
\bibliography{main}

\begin{thebibliography}{24}
\providecommand{\natexlab}[1]{#1}
\providecommand{\url}[1]{\texttt{#1}}
\expandafter\ifx\csname urlstyle\endcsname\relax
  \providecommand{\doi}[1]{doi: #1}\else
  \providecommand{\doi}{doi: \begingroup \urlstyle{rm}\Url}\fi

\bibitem[Adi et~al.(2018)Adi, Baum, Cisse, Pinkas, and Keshet]{adi2018turning}
Yossi Adi, Carsten Baum, Moustapha Cisse, Benny Pinkas, and Joseph Keshet.
\newblock Turning your weakness into a strength: Watermarking deep neural
  networks by backdooring.
\newblock In \emph{27th USENIX security symposium (USENIX Security 18)}, pages
  1615--1631, 2018.

\bibitem[Aiken et~al.(2021)Aiken, Kim, Woo, and Ryoo]{aiken2021neural}
William Aiken, Hyoungshick Kim, Simon Woo, and Jungwoo Ryoo.
\newblock Neural network laundering: Removing black-box backdoor watermarks
  from deep neural networks.
\newblock \emph{Computers \& Security}, 106:\penalty0 102277, 2021.

\bibitem[Bansal et~al.(2022)Bansal, Chiang, Curry, Jain, Wigington, Manjunatha,
  Dickerson, and Goldstein]{bansal2022certified}
Arpit Bansal, Ping-yeh Chiang, Michael~J Curry, Rajiv Jain, Curtis Wigington,
  Varun Manjunatha, John~P Dickerson, and Tom Goldstein.
\newblock Certified neural network watermarks with randomized smoothing.
\newblock In \emph{International Conference on Machine Learning}, pages
  1450--1465. PMLR, 2022.

\bibitem[Fei-Fei et~al.(2006)Fei-Fei, Fergus, and Perona]{fei2006one}
Li~Fei-Fei, Robert Fergus, and Pietro Perona.
\newblock One-shot learning of object categories.
\newblock \emph{IEEE transactions on pattern analysis and machine
  intelligence}, 28\penalty0 (4):\penalty0 594--611, 2006.

\bibitem[Fernandez et~al.(2024)Fernandez, Couairon, Furon, and
  Douze]{fernandez2024functional}
Pierre Fernandez, Guillaume Couairon, Teddy Furon, and Matthijs Douze.
\newblock Functional invariants to watermark large transformers.
\newblock In \emph{ICASSP 2024-2024 IEEE International Conference on Acoustics,
  Speech and Signal Processing (ICASSP)}, pages 4815--4819. IEEE, 2024.

\bibitem[He et~al.(2016)He, Zhang, Ren, and Sun]{he2016deep}
Kaiming He, Xiangyu Zhang, Shaoqing Ren, and Jian Sun.
\newblock Deep residual learning for image recognition.
\newblock In \emph{Proceedings of the IEEE conference on computer vision and
  pattern recognition}, pages 770--778, 2016.

\bibitem[Jain(1989)]{jain1989fundamentals}
Anil~K Jain.
\newblock \emph{Fundamentals of digital image processing}.
\newblock Prentice-Hall, Inc., 1989.

\bibitem[Kirchenbauer et~al.(2023)Kirchenbauer, Geiping, Wen, Katz, Miers, and
  Goldstein]{kirchenbauer2023watermark}
John Kirchenbauer, Jonas Geiping, Yuxin Wen, Jonathan Katz, Ian Miers, and Tom
  Goldstein.
\newblock A watermark for large language models.
\newblock In \emph{International Conference on Machine Learning}, pages
  17061--17084. PMLR, 2023.

\bibitem[Kornblith et~al.(2019)Kornblith, Norouzi, Lee, and
  Hinton]{kornblith2019similarity}
Simon Kornblith, Mohammad Norouzi, Honglak Lee, and Geoffrey Hinton.
\newblock Similarity of neural network representations revisited.
\newblock In \emph{International conference on machine learning}, pages
  3519--3529. PMLR, 2019.

\bibitem[Kreutz-Delgado(2009)]{kreutz2009complex}
Ken Kreutz-Delgado.
\newblock The complex gradient operator and the cr-calculus.
\newblock \emph{arXiv preprint arXiv:0906.4835}, 2009.

\bibitem[Krizhevsky et~al.(2009)Krizhevsky, Hinton,
  et~al.]{krizhevsky2009learning}
Alex Krizhevsky, Geoffrey Hinton, et~al.
\newblock Learning multiple layers of features from tiny images.
\newblock 2009.

\bibitem[Krizhevsky et~al.(2012)Krizhevsky, Sutskever, and
  Hinton]{krizhevsky2012imagenet}
Alex Krizhevsky, Ilya Sutskever, and Geoffrey~E Hinton.
\newblock Imagenet classification with deep convolutional neural networks.
\newblock \emph{Advances in neural information processing systems}, 25, 2012.

\bibitem[Liu et~al.(2021)Liu, Weng, and Zhu]{liu2021watermarking}
Hanwen Liu, Zhenyu Weng, and Yuesheng Zhu.
\newblock Watermarking deep neural networks with greedy residuals.
\newblock In \emph{ICML}, pages 6978--6988, 2021.

\bibitem[Lukas et~al.(2021)Lukas, Zhang, and Kerschbaum]{lukasdeep}
Nils Lukas, Yuxuan Zhang, and Florian Kerschbaum.
\newblock Deep neural network fingerprinting by conferrable adversarial
  examples.
\newblock In \emph{International Conference on Learning Representations}, 2021.

\bibitem[Nematollahi et~al.(2017)Nematollahi, Vorakulpipat, and
  Rosales]{nematollahi2017digital}
Mohammad~Ali Nematollahi, Chalee Vorakulpipat, and Hamurabi~Gamboa Rosales.
\newblock \emph{Digital watermarking}.
\newblock Springer, 2017.

\bibitem[Ren et~al.(2023)Ren, Zhou, Jin, Lyu, and Yan]{ren2023dimension}
Jiaxiang Ren, Yang Zhou, Jiayin Jin, Lingjuan Lyu, and Da~Yan.
\newblock Dimension-independent certified neural network watermarks via
  mollifier smoothing.
\newblock In \emph{International Conference on Machine Learning}, pages
  28976--29008. PMLR, 2023.

\bibitem[Shafieinejad et~al.(2021)Shafieinejad, Lukas, Wang, Li, and
  Kerschbaum]{shafieinejad2021robustness}
Masoumeh Shafieinejad, Nils Lukas, Jiaqi Wang, Xinda Li, and Florian
  Kerschbaum.
\newblock On the robustness of backdoor-based watermarking in deep neural
  networks.
\newblock In \emph{Proceedings of the 2021 ACM workshop on information hiding
  and multimedia security}, pages 177--188, 2021.

\bibitem[Tan et~al.(2023)Tan, Zhong, Qian, Zhang, and Li]{tan2023deep}
Jingxuan Tan, Nan Zhong, Zhenxing Qian, Xinpeng Zhang, and Sheng Li.
\newblock Deep neural network watermarking against model extraction attack.
\newblock In \emph{Proceedings of the 31st ACM International Conference on
  Multimedia}, pages 1588--1597, 2023.

\bibitem[Tang et~al.(2023)Tang, Shen, Zhou, Chen, and Zhang]{tang2023defects}
Ling Tang, Wen Shen, Zhanpeng Zhou, Yuefeng Chen, and Quanshi Zhang.
\newblock Defects of convolutional decoder networks in frequency
  representation.
\newblock In \emph{International Conference on Machine Learning}, pages
  33758--33791. PMLR, 2023.

\bibitem[Uchida et~al.(2017)Uchida, Nagai, Sakazawa, and
  Satoh]{uchida2017embedding}
Yusuke Uchida, Yuki Nagai, Shigeyuki Sakazawa, and Shin'ichi Satoh.
\newblock Embedding watermarks into deep neural networks.
\newblock In \emph{Proceedings of the 2017 ACM on international conference on
  multimedia retrieval}, pages 269--277, 2017.

\bibitem[Wang et~al.(2020)Wang, Wu, Zhang, and Yao]{wang2020watermarking}
Jiangfeng Wang, Hanzhou Wu, Xinpeng Zhang, and Yuwei Yao.
\newblock Watermarking in deep neural networks via error back-propagation.
\newblock \emph{Electronic Imaging}, 32:\penalty0 1--9, 2020.

\bibitem[Xu et~al.(2024)Xu, Wang, Ma, Koh, Xiao, and Chen]{xu2024instructional}
Jiashu Xu, Fei Wang, Mingyu~Derek Ma, Pang~Wei Koh, Chaowei Xiao, and Muhao
  Chen.
\newblock Instructional fingerprinting of large language models.
\newblock \emph{arXiv preprint arXiv:2401.12255}, 2024.

\bibitem[Zeng et~al.(2023)Zeng, Wang, Hu, Xu, Zhou, Wang, Yu, and
  Lin]{zeng2023huref}
Boyi Zeng, Lizheng Wang, Yuncong Hu, Yi~Xu, Chenghu Zhou, Xinbing Wang, Yu~Yu,
  and Zhouhan Lin.
\newblock Huref: Human-readable fingerprint for large language models.
\newblock In \emph{The Thirty-eighth Annual Conference on Neural Information
  Processing Systems}, 2023.

\bibitem[Zhang et~al.(2024)Zhang, Liu, Qian, Zhang, Liu, Qiao, and
  Shao]{zhang2024reef}
Jie Zhang, Dongrui Liu, Chen Qian, Linfeng Zhang, Yong Liu, Yu~Qiao, and Jing
  Shao.
\newblock Reef: Representation encoding fingerprints for large language models.
\newblock \emph{arXiv preprint arXiv:2410.14273}, 2024.

\end{thebibliography}

\newpage
\appendix
\onecolumn

\section{Proofs of our theoretical findings}\label{appendix:theorem}

We first introduce an important equation, which is widely used in the following proofs.

\begin{lemma}\label{le:1}
	\emph{Given $N$ complex numbers, $e^{i n\theta}$, $n=0,1,\ldots,N-1$, the sum of these $N$ complex numbers is given as follows.
		\begin{equation}\label{eq:sum_of_e}
			\forall \theta \in \mathbb{R},\qquad \sum_{n=0}^{N-1}e^{i n\theta} = \frac{\sin(\frac{N \theta}{2})}{\sin(\frac{\theta}{2})}e^{i\frac{(N-1)\theta}{2}} 
		\end{equation}
		Specifically, when $N\theta = 2k\pi , k \in \mathbb{Z}$, $-N < k < N$, we have
		\begin{equation}\label{eq:delta}
			\begin{aligned}
				\forall \theta \in \mathbb{R},\quad &\sum_{n=0}^{N-1}e^{i n\theta} = \frac{\sin(\frac{N \theta}{2})}{\sin(\frac{\theta}{2})}e^{i\frac{(N-1)\theta}{2}} = N \delta_{\theta} ;\quad \text{s.t.} \ N\theta = 2k\pi , k \in \mathbb{Z} ,-N < k < N, \\
				&\textit{where}\quad	\delta_{\theta}=  \begin{cases}
					1,\ &\theta=0 \\ 0,\ & \text{otherwise}\end{cases}
			\end{aligned}
	\end{equation}}
\end{lemma}

We prove Lemma~\ref{le:1} as follows.

\begin{proof}
	First, let us use the letter $S\in \mathbb{C}$ to denote the term of $\sum_{n=0}^{N-1}e^{i n\theta} $.
	\begin{equation*}
		S = \sum_{n=0}^{N-1}e^{i n \theta } 
	\end{equation*}
	
	Therefore, $e^{i\theta }S$ is formulated as follows.
	\begin{equation*}
		e^{i\theta }S = \sum_{n=1}^{N}e^{in\theta } \in \mathbb{C}
	\end{equation*}
	
	Then, $S$ can be computed as $S= \frac{e^{i\theta }S-S}{e^{i\theta }-1}$. Therefore, we have
	\begin{equation*}
		\begin{aligned}
			S & = \frac{e^{i\theta }S-S}{e^{i\theta }-1} \\
			& = \frac{\sum_{n=1}^{N}e^{in\theta } -\sum_{n=0}^{N-1}e^{i n \theta } }{e^{i\theta }-1} \\
			&= \frac{e^{iN\theta } -1 }{e^{i\theta} -1}\\
			&= \frac{e^{i\frac{N\theta }{2}}- e^{-i\frac{N\theta }{2}}}{e^{i\frac{\theta}{2}}- e^{-i\frac{\theta}{2}}} e^{i\frac{(N-1)\theta}{2}} \\
			&= \frac{(e^{i\frac{N\theta }{2}}- e^{-i\frac{N\theta }{2}})/2i}{(e^{i\frac{\theta}{2}}- e^{-i\frac{\theta}{2}})/2i} e^{i\frac{(N-1)\theta}{2}} \\
			&=\frac{\sin(\frac{N \theta}{2})}{\sin(\frac{\theta}{2})}e^{i\frac{(N-1)\theta}{2}}
		\end{aligned}
	\end{equation*}
	Therefore, we prove that $\sum_{n=0}^{N-1}e^{i n \theta } =\frac{\sin(\frac{N \theta}{2})}{\sin(\frac{\theta}{2})}e^{i\frac{(N-1)\theta}{2}}$. 
	
	Then, we prove the special case that when $N\theta = 2k\pi , k \in \mathbb{Z} ,-N < k < N$, $\sum_{n=0}^{N-1}e^{i n\theta} = N \delta_{\theta}=\begin{cases}
		N,\ &\theta=0 \\ 0,\ & \text{otherwise}\end{cases}$, as follows.
	
	When $\theta = 0$, we have
	\begin{equation*}
		\begin{aligned}
			\lim_{\theta \to 0}  \sum_{n=0}^{N-1} e^{in\theta} 
			&= \lim_{\theta \to 0}  \frac{\sin(\frac{N\theta}{2})}{\sin(\frac{\theta}{2})}e^{i\frac{(N-1)\theta}{2}} \\
			&= \lim_{\theta \to 0}  \frac{\sin(\frac{N\theta}{2})}{\sin(\frac{\theta}{2})} \\  
			&= N
		\end{aligned}
	\end{equation*}
	
	When $\theta \ne 0$, and $N\theta = 2k\pi,k \in \mathbb{Z} ,-N < k < N$, we have
	\begin{equation*}
		\begin{aligned}
			\sum_{n=0}^{N-1} e^{in\theta} 
			&= \frac{\sin(\frac{N\theta}{2})}{\sin(\frac{\theta}{2})}e^{i\frac{(N-1)\theta}{2}} \\
			&= \frac{\sin(k\pi)}{\sin(\frac{k\pi}{N})}e^{i\frac{(N-1)k\pi}{N}} \\
			&= 0 \\
		\end{aligned}
	\end{equation*}
	
\end{proof}

In the following proofs, the following two equations are widely used, which are derived based on Lemma~\ref{le:1}.

\begin{equation*}
	\begin{aligned}
		\sum_{m=0}^{M-1}\sum_{n=0}^{N-1} e^{-i(\frac{um}{M}+\frac{vn}{N})2\pi} 
		&= \sum_{m=0}^{M-1}  e^{im(-\frac{u2\pi}{M})} \sum_{n=0}^{N-1}  e^{in(-\frac{v2\pi}{N})}
		\\ 
		& = (M\delta_{-\frac{u2\pi}{M}}) (N\delta_{-\frac{v2\pi}{N}})
		\quad {\rm \slash \slash According\ to\ Equation~(\ref{eq:delta}})\\
		& = \begin{cases}
			MN,\ &u=v=0 \\ 0,\ & \text{otherwise}
		\end{cases}
	\end{aligned}
\end{equation*}

To simplify the representation, \textbf{let $\delta_{uv}$ be the simplification of  $\delta_{-\frac{u2\pi}{M}}\delta_{-\frac{v2\pi}{N}}$ in the following proofs.} Therefore, we have 
\begin{equation}\label{eq:delta_uv}
	\sum_{m=0}^{M-1}\sum_{n=0}^{N-1} e^{-i(\frac{um}{M}+\frac{vn}{N})2\pi} = MN\delta_{uv} = \begin{cases}
		MN,\ &u=v=0 \\ 0,\ & \text{otherwise}
	\end{cases}
\end{equation}

Similarly, we derive the second equation as follows.

\begin{small}\begin{equation}\label{eq:delta_u_v_}
		\begin{aligned}
			\sum_{m=0}^{M-1}\sum_{n=0}^{N-1} e^{i(\frac{(u-u')m}{M}+\frac{(v-v')n}{N})2\pi} 
			&= \sum_{m=0}^{M-1}  e^{im(\frac{(u-u')2\pi}{M})} \sum_{n=0}^{N-1}  e^{in(\frac{(v-v')2\pi}{N})} \\
			& = MN\delta_{\frac{(u-u')2\pi}{M}}\delta_{\frac{(v-v')2\pi}{N}}
			\quad {\rm \slash \slash According\ to\ Equation~(\ref{eq:delta}})\\
			& = MN\delta_{u-u'}\delta_{v-v'}\\
			& = \begin{cases}
				MN,\ &u'=u;v'=v \\ 0,\ & \text{otherwise}
			\end{cases}
		\end{aligned}
\end{equation}\end{small}


\subsection{Proof of Theorem~\ref{th:backward_pro}}\label{app:sec:backward_pro}
In this section, we prove Theorem~\ref{th:backward_pro} in the main paper, as follows.

\begin{proof}
	
	According to the DFT and the inverse DFT, we can obtain the mathematical relationship between $G_{uv}^{(c)}$ and $X_{mn}^{(c)}$, and the mathematical relationship between $Q_{uv}^{(c)}$ and $W_{ts}^{(c)}$, as follows.
	\begin{small}\begin{equation}\label{eq:dft}
			\begin{aligned}
				\left\{
				\begin{aligned}
					& G_{uv}^{(c)} = \sum_{m=0}^{M-1}\sum_{n=0}^{N-1} X_{mn}^{(c)}e^{-i(\frac{um}{M}+\frac{vn}{N})2\pi}  \\
					& X_{mn}^{(c)} = \frac{1}{MN}\sum_{u=0}^{M-1}\sum_{v=0}^{N-1}G_{uv}^{(c)}e^{i(\frac{um}{M}+\frac{vn}{N})2\pi} 
				\end{aligned}
				\right. 
				\left\{
				\begin{aligned}
					& Q_{uv}^{(c)}= \sum_{t=0}^{K-1}\sum_{s=0}^{K-1} W_{ts}^{(c)}e^{i(\frac{ut}{M}+\frac{vs}{N})2\pi} \\
					& W_{ts}^{(c)} = \frac{1}{MN}\sum_{u=0}^{M-1}\sum_{v=0}^{N-1}Q_{uv}^{(c)}e^{-i(\frac{ut}{M}+\frac{vs}{N})2\pi} 
				\end{aligned}
				\right.
			\end{aligned}
	\end{equation}\end{small}
	
	Based on Equation~(\ref{eq:dft}) and the derivation rule for complex numbers \citep{kreutz2009complex}, we can obtain the mathematical relationship between $\frac{\partial \textit{Loss}}{\partial \overline{G}_{uv}^{(c)}}$ and $\frac{\partial \textit{Loss}}{\partial \overline{X}_{mn}^{(c)}}$, and the mathematical relationship between $\frac{\partial \textit{Loss}}{\partial  \overline{Q}_{uv}^{(c)}}$ and $\frac{\partial \textit{Loss}}{\partial \overline{W}_{ts}^{(c)} }$, as follows. Note that when we use gradient descent to optimize a real-valued loss function $\textit{Loss}$ with complex variables, people usually treat the real and imaginary values, $a\in\mathbb{C}$ and $b\in\mathbb{C}$, of a complex variable ($z=a+bi$) as two separate real-valued variables, and separately update these two real-valued variables. In this way, the exact optimization step of $z$ computed based on such a technology is equivalent to $\frac{\partial \textit{Loss}}{\partial \overline{z}}$. Since $X_{mn}^{(c)}$ and $W_{cts}^{(l)[\text{ker=d}]}$ are real numbers, $\frac{\partial Loss}{\partial \overline{X}^{(c)}_{mn}} =	\frac{\partial Loss}{\partial X_{mn}^{(c)}}$ and $\frac{\partial Loss}{\partial \overline{W}_{ts}^{(c)}} =	\frac{\partial Loss}{\partial W_{ts}^{(c)}}$.
	
	\begin{small}\begin{equation}\label{eq:dft_grad}
			\begin{aligned}
				\left\{
				\begin{aligned}
					& \frac{\partial Loss}{\partial \overline{G}^{(c)}_{uv}} =\frac{1}{MN}\sum_{m=0}^{M-1}\sum_{n=0}^{N-1}\frac{\partial Loss}{\partial \overline{X}^{(c)}_{mn}}e^{-i(\frac{um}{M}+\frac{vn}{N})2\pi}  \\
					& \frac{\partial Loss}{\partial \overline{X}^{(c)}_{mn}} =\sum_{u=0}^{M-1}\sum_{v=0}^{N-1} \frac{\partial Loss}{\partial \overline{G}^{(c)}_{uv}}e^{i(\frac{um}{M}+\frac{vn}{N})2\pi}
				\end{aligned}
				\right.
				\left\{
				\begin{aligned}
					& \frac{\partial Loss}{\partial \overline{Q}^{(c)}_{uv}}= \frac{1}{MN}\sum_{t=0}^{K-1}\sum_{s=0}^{K-1} \frac{\partial Loss}{\partial  \overline{W}^{(c)}_{ts}}e^{i(\frac{ut}{M}+\frac{vs}{N})2\pi}  \\
					& \frac{\partial Loss}{\partial \overline{W}^{(c)}_{ts}} = \sum_{u=0}^{M-1}\sum_{v=0}^{N-1}\frac{\partial Loss}{\partial \overline{Q}^{(c)}_{uv}}e^{-i(\frac{ut}{M}+\frac{vs}{N})2\pi} 
				\end{aligned}
				\right.
			\end{aligned}
	\end{equation}\end{small}

	Let us conduct the convolution operation on the feature map $\mathbf{X}=[X^{(1)},X^{(2)},\cdots,X^{(C)}]\in\mathbb{R}^{C\times M\times N}$, and obtain the output feature map $Y\in\mathbb{R}^{M\times N}$ as follows.
	\begin{equation}\label{app_eq:conv}
		Y_{mn} =b + \sum_{c=1}^{C} \sum_{t=0}^{K-1}\sum_{s=0}^{K-1}W_{ts}^{(c)} X^{(c)}_{m+t,n+s}\\ 
	\end{equation}
	
	Based on Equation~(\ref{eq:dft}) and Equation~(\ref{eq:dft_grad}), and the derivation rule for complex numbers \citep{kreutz2009complex}, the exact optimization step of $Q_{uv}^{(c)}$ in real implementations can be computed as follows.
	\begin{small}\begin{equation*}
			\begin{aligned}
				&\frac{\partial \textit{Loss}}{\partial \overline{Q}^{(c)}_{uv}} \\
				&= \frac{1}{MN}\sum_{t=0}^{K-1}\sum_{s=0}^{K-1} \frac{\partial \textit{Loss}}{\partial \overline{W}_{ts}^{(c)}}e^{i(\frac{ut}{M}+\frac{vs}{N})2\pi} 
				\quad {\rm \slash \slash Equation~(\ref{eq:dft_grad})}\\
				& =\frac{1}{MN}\sum_{t=0}^{K-1}\sum_{s=0}^{K-1} \left( \sum_{m=0}^{M-1}\sum_{n=0}^{N-1} \frac{\partial \textit{Loss}}{\partial \overline{Y}_{mn}}\cdot \overline{X}^{(c)}_{m+t,n+s}\right)e^{i(\frac{ut}{M}+\frac{vs}{N})2\pi} 
				\quad {\rm \slash \slash Equation~(\ref{app_eq:conv})}\\
				&  \quad {\rm \slash \slash Equation~(\ref{eq:dft})} \\
				& = \frac{1}{MN}\sum_{t=0}^{K-1}\sum_{s=0}^{K-1} \left(\sum_{m=0}^{M-1}\sum_{n=0}^{N-1} \frac{\partial \textit{Loss}}{\partial \overline{Y}_{mn}}\cdot  \frac{1}{MN}\sum_{u'=0}^{M-1}\sum_{v'=0}^{N-1} \overline{G}^{(c)}_{u'v'} e^{-i(\frac{u'(m+t)}{M}+\frac{v'(n+s)}{N})2\pi} \right)e^{i(\frac{ut}{M}+\frac{vs}{N})2\pi}\\
				& = \frac{1}{MN}\sum_{t=0}^{K-1}\sum_{s=0}^{K-1} \left( \sum_{u'=0}^{M-1}\sum_{v'=0}^{N-1} \overline{G}^{(c)}_{u'v'} e^{-i(\frac{u't}{M}+\frac{v's}{N})2\pi} \cdot \frac{1}{MN}\sum_{m=0}^{M-1}\sum_{n=0}^{N-1}\frac{\partial \textit{Loss}}{\partial \overline{Y}_{mn}} e^{-i(\frac{u'm}{M}+\frac{v'n}{N})2\pi} \right)e^{i(\frac{ut}{M}+\frac{vs}{N})2\pi} \\
				& = \frac{1}{MN}\sum_{t=0}^{K-1}\sum_{s=0}^{K-1} \left( \sum_{u'=0}^{M-1}\sum_{v'=0}^{N-1}\overline{G}^{(c)}_{u'v'}  \frac{\partial Loss}{\partial \overline{H}_{u'v'}}e^{-i(\frac{u't}{M}+\frac{v's}{N})2\pi} \right)e^{i(\frac{ut}{M}+\frac{vs}{N})2\pi} \quad {\rm \slash \slash Equation~(\ref{eq:dft_grad})} \\
				&= \frac{1}{MN}\sum_{t=0}^{K-1}\sum_{s=0}^{K-1} \sum_{u'=0}^{M-1}\sum_{v'=0}^{N-1}\overline{G}^{(c)}_{u'v'}  \frac{\partial \textit{Loss}}{\partial \overline{H}_{u'v'}} e^{i(\frac{(u-u')t}{M}+\frac{(v-v')s}{N})2\pi}\\
				&=  \sum_{u'=0}^{M-1}\sum_{v'=0}^{N-1}\overline{G}^{(c)}_{u'v'}  \frac{\partial \textit{Loss}}{\partial \overline{H}_{u'v'}} \cdot \frac{1}{MN}\sum_{t=0}^{K-1}\sum_{s=0}^{K-1}  e^{i(\frac{(u-u')t}{M}+\frac{(v-v')s}{N})2\pi} \\
				& \slash \slash \text{ Let} \ \ A_{u'v'uv} = \sum_{t=0}^{K-1}\sum_{s=0}^{K-1}  e^{i(\frac{(u-u')t}{M}+\frac{(v-v')s}{N})2\pi} \\
				&= \frac{1}{MN} \sum_{u'=0}^{M-1} \sum_{v'=0}^{N-1}  \overline{G}^{(c)}_{u'v'}  \frac{\partial \textit{Loss}}{\partial \overline{H}_{u'v'}} 
			\end{aligned}
	\end{equation*}\end{small}
	where $A_{u'v'uv}$ can be rewritten as follows.
	\begin{small}\begin{equation*}
			\begin{aligned}
				A_{u'v'uv}
				&= \sum_{t=0}^{K-1}\sum_{s=0}^{K-1}  e^{i(\frac{(u-u')t}{M}+\frac{(v-v')s}{N})2\pi}\\
				&= \sum_{t=0}^{K-1}e^{i\frac{(u-u')2\pi}{M} t} \sum_{s=0}^{K-1} e^{i\frac{(v-v')2\pi}{N} s}\\
				&=   \frac{\sin(\frac{K(u-u')\pi}{M})}{\sin(\frac{(u-u')\pi}{M})}  \frac{\sin(\frac{K(v-v')\pi}{N})}{\sin(\frac{(v-v')\pi}{N})} \cdot e^{i(\frac{(K-1)(u-u')}{M}+\frac{(K-1)(v-v')}{N})\pi}  \quad {\rm \slash \slash According\ to\ Equation\ (\ref{eq:sum_of_e})} \\
			\end{aligned}
	\end{equation*}\end{small}

	Based on the derived $\frac{\partial \textit{Loss}}{\partial \overline{Q}^{(c)}_{uv}} \in\mathbb{C}$ , we can further compute gradients $\frac{\partial Loss}{\partial (\overline{T}^{(l,uv)})^{\top}}\in\mathbb{C}^{D\times C}$ as follows.
	\begin{small}\begin{equation}\label{eq:bp_grad_single_T}
			\frac{\partial Loss}{\partial \overline{\mathcal{F_{\mathbf{W}}}}^{(uv)}} =  \frac{1}{MN}\sum\limits_{u'=0}^{M-1}\sum\limits_{v'=0}^{N-1}A_{uvu'v'} \frac{\partial \textit{Loss}}{\partial \overline{\mathcal{F}}_{Y}^{(u'v')}} \cdot  \overline{\mathcal{F}}_{\mathbf{X}}^{(u'v')}
	\end{equation}\end{small}
	
	Let us use the gradient descent algorithm to update the convlutional weight $W_{ts}^{(c)}|_{n}$ of the $n$-th epoch, the updated frequency spectrum $W_{ts}^{(c)}|_{n+1}$ can be computed as follows.
	
	\begin{equation*}
		\forall t,s,\quad W_{ts}^{(c)}|_{n+1} = W_{ts}^{(c)}|_{n} - \eta \cdot \frac{\partial Loss}{\partial \overline{W}_{ts}^{(c)}}
	\end{equation*}
	
	where $\eta$ is the learning rate. Then, the updated frequency spectrum $T^{(l,uv)}|_{n+1}$ computed based on Equation~(\ref{eq:dft_grad}) is given as follows.
	
	\begin{equation*}
		\begin{aligned}
			\Delta Q_{uv}^{(c)} 
			&= Q_{uv}^{(c)}|_{n+1} - Q_{uv}^{(c)}|_{n} \\
			&=  \sum_{t=0}^{K-1}\sum_{s=0}^{K-1} W_{ts}^{(c)}|_{n+1}e^{i(\frac{ut}{M}+\frac{vs}{N})2\pi} - Q_{uv}^{(c)}|_{n}
			\quad {\rm \slash \slash Equation~(\ref{eq:dft})}\\
			&= \sum_{t=0}^{K-1}\sum_{s=0}^{K-1} (W_{ts}^{(c)}|_{n} - \eta \cdot \frac{\partial Loss}{\partial \overline{W}_{ts}^{(c)}})  e^{i(\frac{ut}{M}+\frac{vs}{N})2\pi} - Q_{uv}^{(c)}|_{n} \\
			&= (\sum_{t=0}^{K-1}\sum_{s=0}^{K-1} W_{ts}^{(c)}|_{n} e^{i(\frac{ut}{M}+\frac{vs}{N})2\pi}- Q_{uv}^{(c)}|_{n}) - \eta \sum_{t=0}^{K-1}\sum_{s=0}^{K-1} \frac{\partial Loss}{\partial \overline{W}_{ts}^{(c)}}  e^{i(\frac{ut}{M}+\frac{vs}{N})2\pi} \\
			&=  - \eta \sum_{t=0}^{K-1}\sum_{s=0}^{K-1} \frac{\partial Loss}{\partial \overline{W}_{ts}^{(c)}} e^{i(\frac{ut}{M}+\frac{vs}{N})2\pi} 
			\quad {\rm \slash \slash Equation~(\ref{eq:dft})}\\
			&= - \eta MN \frac{\partial Loss}{\partial \overline{Q}_{uv}^{(c)}}
			\quad {\rm \slash \slash Equation~(\ref{eq:dft_grad})}\\
		\end{aligned}
	\end{equation*}
	
	Therefore, we prove that any step on $W^{(c)}_{ts}$  equals to $MN$ step on $Q_{uv}^{(c)}$. In this way, the change of frequency components $\mathcal{F}^{(uv)}_{\mathbf{W}}$ can be computed as follows.
	\begin{small}
		\begin{equation}
			\Delta \mathcal{F}^{(uv)}_{\mathbf{W}} =  - \eta 
        \sum\limits_{u'=0}^{M-1}\sum\limits_{v'=0}^{N-1}A_{uvu'v'} \frac{\partial \textit{Loss}}{\partial \overline{\mathcal{F}}_{Y}^{(u'v')}} \cdot  \overline{\mathcal{F}}_{\mathbf{X}}^{(u'v')}
		\end{equation}
	\end{small}
 \end{proof}

 \subsection{Proof of Corollary~\ref{co:zero}}\label{app:sec:zero}
In this section, we prove Corollary~\ref{co:zero} in Section~\ref{sec:meth} of the main paper, as follows.

\begin{proof}
According to Theorem~\ref{th:backward_pro}, the change of the frequency components at frequencies $(u,v) \in S$  can be further derived as follows.

\begin{small}\begin{equation*}
        \begin{aligned}
        \Delta \mathcal{F}^{(uv)}_{\mathbf{W}} &=  - \eta 
        \sum\limits_{u'=0}^{M-1}\sum\limits_{v'=0}^{N-1}A_{uvu'v'} \frac{\partial \textit{Loss}}{\partial \overline{\mathcal{F}}_{Y}^{(u'v')}} \cdot  \overline{\mathcal{F}}_{\mathbf{X}}^{(u'v')} \\
        & = - \eta A_{uv00} \frac{\partial \textit{Loss}}{\partial \overline{\mathcal{F}}_{Y}^{(00)}} \cdot  \overline{\mathcal{F}}_{\mathbf{X}}^{(00)} \quad {\rm \slash \slash \forall(u,v)\ne (0,0), \mathcal{F}_{\mathbf{X}}^{(uv)}=0}\\
        &= - \eta \frac{\partial \textit{Loss}}{\partial \overline{\mathcal{F}}_{Y}^{(00)}} \cdot  \overline{\mathcal{F}}_{\mathbf{X}}^{(00)} \cdot
\frac{\sin(\frac{Ku\pi}{M})}{\sin(\frac{u\pi}{M})}  \frac{\sin(\frac{Kv\pi}{N})}{\sin(\frac{v\pi}{N})} \cdot e^{i(\frac{(K-1)u}{M}+\frac{(K-1)v}{N})\pi}\\
&= - \eta \frac{\partial \textit{Loss}}{\partial \overline{\mathcal{F}}_{Y}^{(00)}} \cdot  \overline{\mathcal{F}}_{\mathbf{X}}^{(00)} \cdot e^{i(\frac{(K-1)i\pi}{K}+\frac{(K-1)j\pi}{K})\pi} \cdot
\frac{\sin(i\pi)}{\sin(i\pi/K)}  \frac{\sin(j\pi)}{\sin(j\pi/K)}  \\ & {\rm \slash \slash S = \{ (u, v) \ | \ u = iM/K \ \text{or} \ v = jN/K; \ i, j \in \{1, 2, \dots, K-1\} \}} \\
&= 0 \quad {\rm \slash \slash \sin(i\pi)=0} \
        \end{aligned} 
\end{equation*}\end{small}

Therefore, we have proved the frequency components {\small $\mathcal{F}_{\mathbf{W}}^{(uv)}$} at the frequencies in the set {\small $S$} keep invariant.
    
\end{proof}

 \subsection{Proof of Theorem~\ref{th:scale}}\label{app:sec:scale}
In this section, we prove Theorem~\ref{th:scale} in Section~\ref{sec:meth} of the main paper, as follows.

\begin{proof}
    The frequency components $\mathcal{F}_{\mathbf{W}^*}^{(uv)}$ of the scaled filter $\mathbf{W}^* = a \cdot \mathbf{W}$ are computed as follows.
    \begin{small}\begin{equation*}
    \begin{aligned}
        \mathcal{F}_{\mathbf{W}^*}^{(uv)}  &=  [Q_{uv}^{*(1)}, Q_{uv}^{*(2)}, \ldots, Q_{uv}^{*(C)}]^\top \\
        &= [\sum_{t=0}^{K-1} \sum_{s=0}^{K-1} W_{ts}^{*(1)} e^{i(\frac{ut}{M} + \frac{vs}{N}) 2\pi},\sum_{t=0}^{K-1} \sum_{s=0}^{K-1} W_{ts}^{*(2)} e^{i(\frac{ut}{M} + \frac{vs}{N}) 2\pi},\cdots,\sum_{t=0}^{K-1} \sum_{s=0}^{K-1} W_{ts}^{*(C)} e^{i(\frac{ut}{M} + \frac{vs}{N}) 2\pi}]  \\
        &= [\sum_{t=0}^{K-1} \sum_{s=0}^{K-1} a \cdot W_{ts}^{(1)} e^{i(\frac{ut}{M} + \frac{vs}{N}) 2\pi},\sum_{t=0}^{K-1} \sum_{s=0}^{K-1} a \cdot W_{ts}^{(2)} e^{i(\frac{ut}{M} + \frac{vs}{N}) 2\pi},\cdots,\sum_{t=0}^{K-1} \sum_{s=0}^{K-1} a \cdot W_{ts}^{(C)} e^{i(\frac{ut}{M} + \frac{vs}{N}) 2\pi}]^\top \\
        &= a \cdot [\sum_{t=0}^{K-1} \sum_{s=0}^{K-1}  W_{ts}^{(1)} e^{i(\frac{ut}{M} + \frac{vs}{N}) 2\pi},\sum_{t=0}^{K-1} \sum_{s=0}^{K-1} W_{ts}^{(2)} e^{i(\frac{ut}{M} + \frac{vs}{N}) 2\pi},\cdots,\sum_{t=0}^{K-1} \sum_{s=0}^{K-1}  W_{ts}^{(C)} e^{i(\frac{ut}{M} + \frac{vs}{N}) 2\pi}]^\top \\
        & = a \cdot [Q_{uv}^{(1)}, Q_{uv}^{(2)}, \ldots, Q_{uv}^{(C)}]^\top \\
        &= a \cdot \mathcal{F}_{\mathbf{W}}^{(uv)}
    \end{aligned}
    \end{equation*}\end{small}

Thus,  we have proved that the frequency components $\mathcal{F}_{\mathbf{W}^*}^{(uv)}$ of the scaled filter are equal to the scaled frequency components $a\cdot \mathcal{F}_{\mathbf{W}}^{(uv)}$ of the original filter.
    
\end{proof}

 \subsection{Proof of Theorem~\ref{th:permu}}\label{app:sec:permu}
In this section, we prove Theorem~\ref{th:permu} in Section~\ref{sec:meth} of the main paper, as follows.

\begin{proof}
    The frequency components $[\mathcal{F}^{(uv)}_{\mathbf{W}_{\pi(1)}},\cdots, \mathcal{F}^{(uv)}_{\mathbf{W}_{\pi(D)}} ]$ of the permuted filters $[\mathbf{W}_{\pi(1)},\mathbf{W}_{\pi(2)},\cdots, \mathbf{W}_{\pi(D)} ]$ are computed as follows. 

    \begin{small}
        \begin{equation}
        \begin{aligned}
        \left [\mathcal{F}^{(uv)}_{\mathbf{W}_{\pi(1)}},\cdots, \mathcal{F}^{(uv)}_{\mathbf{W}_{\pi(D)}} \right ] 
        &= \left [\mathcal{T}_{uv}(\mathbf{W}_{\pi(1)}),\mathcal{T}_{uv}(\mathbf{W}_{\pi(2)}),\cdots, \mathcal{T}_{uv}(\mathbf{W}_{\pi(D)})\right ]\quad {\rm \slash \slash Equation~(\ref{eq:com_filter})}\\ 
        &= \pi \left [\mathcal{T}_{uv}(\mathbf{W}_{1}),\mathcal{T}_{uv}(\mathbf{W}_{2}),\cdots, \mathcal{T}_{uv}(\mathbf{W}_{D})\right ]\\
        &= \pi \left [\mathcal{F}^{(uv)}_{\mathbf{W}_{1}},\cdots, \mathcal{F}^{(uv)}_{\mathbf{W}_{D}} \right ]
        \end{aligned}
        \end{equation}
    \end{small}

    The frequency components $[\mathcal{F}^{(uv)}_{\mathbf{W}_{\pi(1)}},\cdots, \mathcal{F}^{(uv)}_{\mathbf{W}_{\pi(D)}} ]$ of the permuted filters $[\mathbf{W}_{\pi(1)},\mathbf{W}_{\pi(2)},\cdots, \mathbf{W}_{\pi(D)} ]$ are equal to the permuated frequency components $\pi [\mathcal{F}^{(uv)}_{\mathbf{W}_{1}},\cdots, \mathcal{F}^{(uv)}_{\mathbf{W}_{D}} ]$.

\end{proof}


\end{document}